%% file: main.tex
\documentclass{article}
\usepackage{iclr2027_conference,times}
\iclrpreprintcopy

\input{math_commands.tex}

\usepackage{hyperref,url,graphicx,booktabs,amsmath,amssymb,array,multirow,enumitem,xcolor,colortbl,microtype,fvextra,longtable,float,placeins}
\input{author_info.tex}
\hypersetup{hidelinks,hypertexnames=false,pdfauthor={\DSABenchPDFAuthors},pdftitle={DSABench: Benchmarking Source Discovery, Coordination, and Execution for Multi-Database Agents}}
\title{DSABench: Benchmarking Source Discovery,\\Coordination, and Execution for\\Multi-Database Agents}
\author{\DSABenchAuthorBlock}
\begin{document}
\maketitle
\begin{abstract}
Most text-to-SQL benchmarks provide the target database. In a large data catalog, however, choosing the sources is part of the problem: an agent must find every database needed for a request, determine how their results fit together, and execute the resulting workflow. We present \textbf{DSABench} to evaluate these capabilities jointly. Its 815 human-reviewed tasks draw on Spider, BIRD, and LiveSQLBench; 618 require several databases to contribute, and 105 allow more than one interpretation of the required sources. Executable contracts define source participation, cross-source dependencies, and source-local execution, while \textbf{RouteEnv} lets agents inspect public catalog evidence without revealing the gold annotations. In our routing-only study across seven model backends, the highest \textbf{RA-Set} score is \textbf{60.55\%}, compared with a highest \textbf{RA-Graph} score of \textbf{34.66\%} when coordination requirements are included. The gap becomes more pronounced when execution is required. Across four backends, median non-privileged \textbf{Route+Exec-Graph} success ranges from \textbf{6.69\% to 7.98\%} without clarification and from \textbf{11.44\% to 13.25\%} with one clarification; the best observed clarified configuration reaches \textbf{16.99\%}. On ambiguous tasks, clarification improves exact branch-source recovery by \textbf{39.8--43.2 percentage points}. These findings illustrate why retrieving plausible databases or producing correct local SQL is not enough: an agent must carry the right information through the entire multi-database workflow.

\end{abstract}

\section{Introduction}
Most text-to-SQL evaluations start with the database already selected. Spider tests generalization to unseen schemas, BIRD emphasizes database grounding and external knowledge, and newer benchmarks examine enterprise workflows, robustness, ambiguity, and interaction~\citep{yu2018spider,li2023bird,lei2025spider2,chang2023drspider,dong2025practiq,huo2026birdinteract,livesqlbench2025}. Yet an agent working with a large catalog cannot take the choice of database for granted. It must work out which sources are needed, whether one query depends on the output of another, and whether the individual queries can be executed successfully. These decisions form the source discovery, coordination, and execution problem that DSABench evaluates.

Consider the request in Figure~\ref{fig:workflow}: count the vocal types for a song whose title matches that of the newest movie. One database identifies the movie; another contains the song and vocal records. Querying only the movie database cannot answer the request. Even selecting both databases is insufficient unless the movie title returned by the first query is passed to the second. The agent must do more than identify relevant schemas or write syntactically valid SQL. It must put the queries together in the right order, with the right information flowing between them.

Prior work on database discovery includes DBCopilot, DBRouting, LinkAlign, Bi-SR, and MDB-Link~\citep{wang2025dbcopilot,mandal2025dbrouting,wang2025linkalign,wu2026bisr,xu2026mdblink}. DSABench asks a different evaluation question: \textbf{did the agent identify and use every source required by the task?} Some requests need several independent databases; others need the result of one query to parameterize another. A catalog relation may suggest that two sources are compatible, but it cannot establish that a particular request requires data to flow between them. Nor is returning a list of plausible candidates the same as recovering the complete required set. Ambiguity adds another layer: after clarification, even the identity of the required source may change.

We build DSABench from 815 human-reviewed tasks based on Spider, BIRD, and LiveSQLBench: 618 require multiple databases jointly, 92 have a single required source, and 105 admit alternative source interpretations. Candidate catalogs range from 11 to 166 databases. We begin with executable source contributions and explicit coordination contracts, render them into natural-language requests, and then apply deterministic validation, human review, and adjudication. This process supplies a checkable target for source selection, structure, and execution. It does not assume that executable gold queries, by themselves, guarantee that every constraint is clear in the natural-language request.

Agents also need a way to investigate the catalog without seeing the answers. \textbf{RouteEnv} provides public search and inspection tools but withholds the annotated gold sources and SQL. Agents can use that evidence to propose and revise a route; \textbf{Hermes} serves as one reference policy rather than a mandatory part of the benchmark. We evaluate whether the route recovers the complete source set and required structure, whether each source contributes a correct execution result, and whether these conditions hold together. For ambiguous tasks, we score both possible interpretations with equal weight while retaining one task in the 815-task denominator.

The results reveal how much performance is lost between identifying sources and finishing a workflow. In the routing-only study across seven model backends, the highest exact source-set score is 60.55\% (RA-Set), whereas the highest structure-aware score is 34.66\% (RA-Graph). In the non-privileged routing-and-execution study, median Route+Exec-Graph success across four backends is 6.69\%--7.98\% without clarification and 11.44\%--13.25\% with one clarification. Even the best observed clarified configuration succeeds on only 16.99\% of task weight. The drop is not confined to source selection. In a fixed non-privileged setting, 55.1\%--58.4\% of source-correct task weight also meets the structure requirements, and 32.3\%--37.0\% of structure-correct weight passes joint execution. Providing router-generated local requests improves source-level execution for all four backends, but its effect on complete task success is inconsistent. On the 105 ambiguous tasks, a matched one-clarification comparison improves exact branch-source recovery by 39.8--43.2 percentage points. Taken together, these results show why each stage must be measured alongside the complete workflow.

We make three contributions. First, \textbf{DSABench} brings source discovery, cross-database coordination, and source-local execution into a single 815-task evaluation. Second, \textbf{RouteEnv} provides a public catalog interface while keeping the source, structure, and execution targets private to the evaluator. Third, we report both intermediate and end-to-end results, with controlled analyses of local request formulation and clarification. This makes it possible to see not just where agents improve, but whether those improvements help them complete the task.

\section{Related Work}
\paragraph{Database-given and agentic Text-to-SQL.}
Spider and BIRD focus primarily on generating SQL against a specified database~\citep{yu2018spider,li2023bird}. Spider 2.0 and LiveSQLBench introduce more realistic workflows and execution settings; Dr.Spider, PRACTIQ, and BIRD-Interact study robustness, ambiguity, or interaction~\citep{lei2025spider2,livesqlbench2025,chang2023drspider,dong2025practiq,huo2026birdinteract}. SParC and CoSQL investigate context-dependent and conversational SQL~\citep{yu2019sparc,yu2019cosql}, but their dialogue context is different from passing values between databases. RAT-SQL and RESDSQL address schema linking, PICARD constrains decoding, and DIN-SQL, CHESS, and CHASE-SQL use decomposition or candidate refinement~\citep{wang2020ratsql,li2023resdsql,scholak2021picard,pourreza2023dinsql,talaei2024chess,pourreza2025chasesql}. DSABench introduces an earlier decision into the evaluation: which combination of databases must participate in the answer?

\paragraph{Database and schema routing.}
DBRouting ranks databases by their ability to answer a question, although its released evaluation assigns each question one gold database; DBCopilot identifies a database and its relevant tables~\citep{mandal2025dbrouting,wang2025dbcopilot}. LinkAlign and MDB-Link locate a target database and useful schema context, while Bi-SR studies large-scale schema retrieval~\citep{wang2025linkalign,xu2026mdblink,wu2026bisr}. In contrast, DSABench evaluates the exact participating set, the way those sources must coordinate, and their local execution results (Table~\ref{tab:related_scope}). We use RouteEnv-compatible reference policies in our experiments; we do not claim to reproduce or outperform these specialized systems.

\begin{table}[t]
\caption{Primary evaluated object in the nearest database-routing work. The comparison describes the cited evaluation target, not the maximum capability of an extendable system.}
\label{tab:related_scope}
\centering\small
\setlength{\tabcolsep}{4.2pt}
\begin{tabular}{@{}p{0.22\linewidth}p{0.69\linewidth}@{}}
\toprule
Work & Primary routing / evaluation object \\
\midrule
DBRouting & database answerability / ranking for a single gold database \\
DBCopilot & target-database routing together with relevant-table selection \\
LinkAlign / Bi-SR / MDB-Link & database and schema-item localization for SQL generation \\
\textbf{DSABench} & \textbf{exact database set, cross-database route structure, and source-local execution} \\
\bottomrule
\end{tabular}
\end{table}

\paragraph{Agents and evaluation validity.}
ReAct combines reasoning with action, and ToolLLM investigates API use~\citep{yao2023react,qin2024toolllm}. AgentBench, GAIA, and DSBench assess agent task performance; Data Interpreter proposes an agent system~\citep{liu2024agentbench,mialon2024gaia,jing2025dsbench,hong2025datainterpreter}. Binder incorporates model calls into symbolic programs~\citep{cheng2023binder}. Our concern is more specific: a value returned by one database query may have to be supplied to another. Distributed query engines can execute plans over known data sources~\citep{begoli2018calcite}, whereas DSABench evaluates whether an agent can identify the sources and infer the necessary connections in the first place. Because evaluation also depends on how tasks are defined and SQL results are compared~\citep{finegan2018evaluation,zhong2020testsuite}, we state our corpus-specific correctness rules rather than treating execution scores as interchangeable with upstream leaderboard results.

\section{Defining and Constructing the Task}
\label{sec:benchmark}
Each task consists of a user request $q$, a candidate catalog $\mathcal C$, and public metadata $M$. Hidden evaluator annotations specify the required source set $G\subseteq\mathcal C$, a behavior label $\tau$, directed dependencies $E$, and the source-local queries needed to complete the task. The system predicts selected sources $P$, topology $\hat\tau$, dependencies $\hat E$, local requests $\hat Q$, and source SQL $\hat S$. For ordinary tasks, $G$ is the complete set of jointly required databases. Interactive tasks instead specify alternative sources, and each episode is judged against its latent branch target $G_b$. Routers cannot see the gold annotations, SQL, or behavior labels. Importantly, the catalog of eligible databases is not the same as a retrieved Top-$K$ list: the latter is just one observation from the agent's evidence-gathering process.

\begin{figure}[t]
\centering
\includegraphics[width=\linewidth]{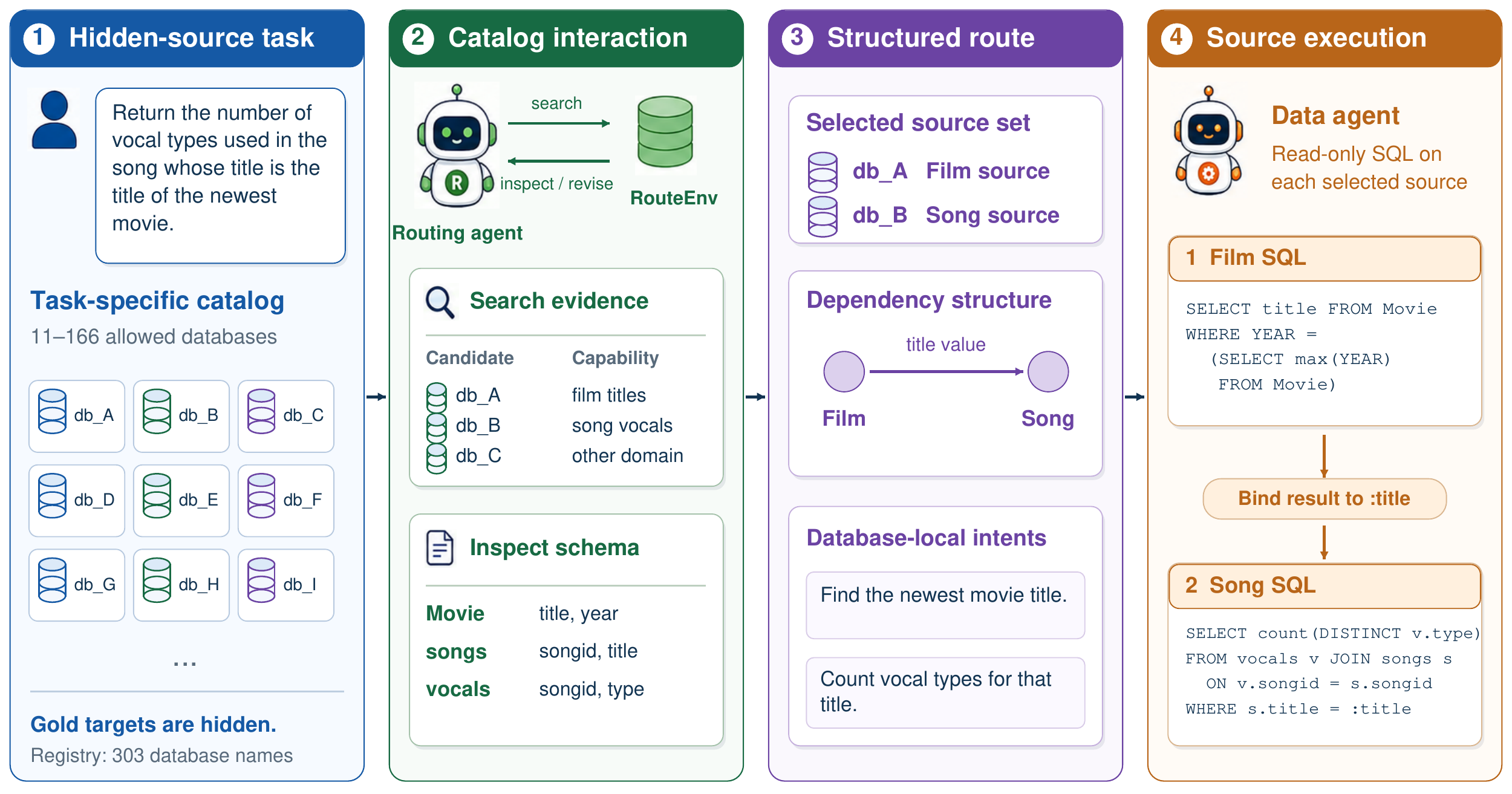}
\caption{Schematic hidden-source sequential workflow (not a frozen benchmark record). The agent searches and inspects public catalog evidence, predicts both sources and their title dependency, then supplies local requests to SQL executors. The request, schemas, and SQL illustrate the output-to-input dependency; source aliases are schematic. Gold targets and SQL remain evaluator-only.}
\label{fig:workflow}
\end{figure}

\paragraph{Five regimes of source uncertainty.}
\textbf{Single-target} tasks require one database. \textbf{Parallel} tasks require independent contributions from several databases, while \textbf{Sequential} tasks pass an upstream query result into a downstream query. \textbf{Exploratory} tasks ask the agent to find sources whose capabilities collectively cover a request. \textbf{Interactive} tasks contain ambiguity that must be resolved between alternative source interpretations. Thus the five regimes stress different aspects of the same problem: identifying sources, finding all of them, recovering data dependencies, matching capabilities, and interpreting user intent. All 815 tasks contribute to the main aggregate; Interactive tasks are evaluated branch by branch in a controlled one-turn setting (Appendix~\ref{app:validity}).

\paragraph{Contract-first construction.}
We build tasks around required executable contributions, not merely databases that happen to be relevant. Each source unit has an owner database, a natural-language intent, read-only gold SQL, and verification evidence. We combine source units only within the same upstream corpus. Before a task is worded in natural language, its contract specifies the sources, local queries, topology, dependencies, value bindings, and any applicable branch rules. A language renderer expresses that contract as a user request. Deterministic checks then verify database ownership, schema references, read-only execution, source requirements, and structural consistency. These checks establish that the task is grounded in executable data; they do not establish that its wording expresses every gold constraint unambiguously.

\begin{figure}[t]
\centering
\includegraphics[width=\linewidth]{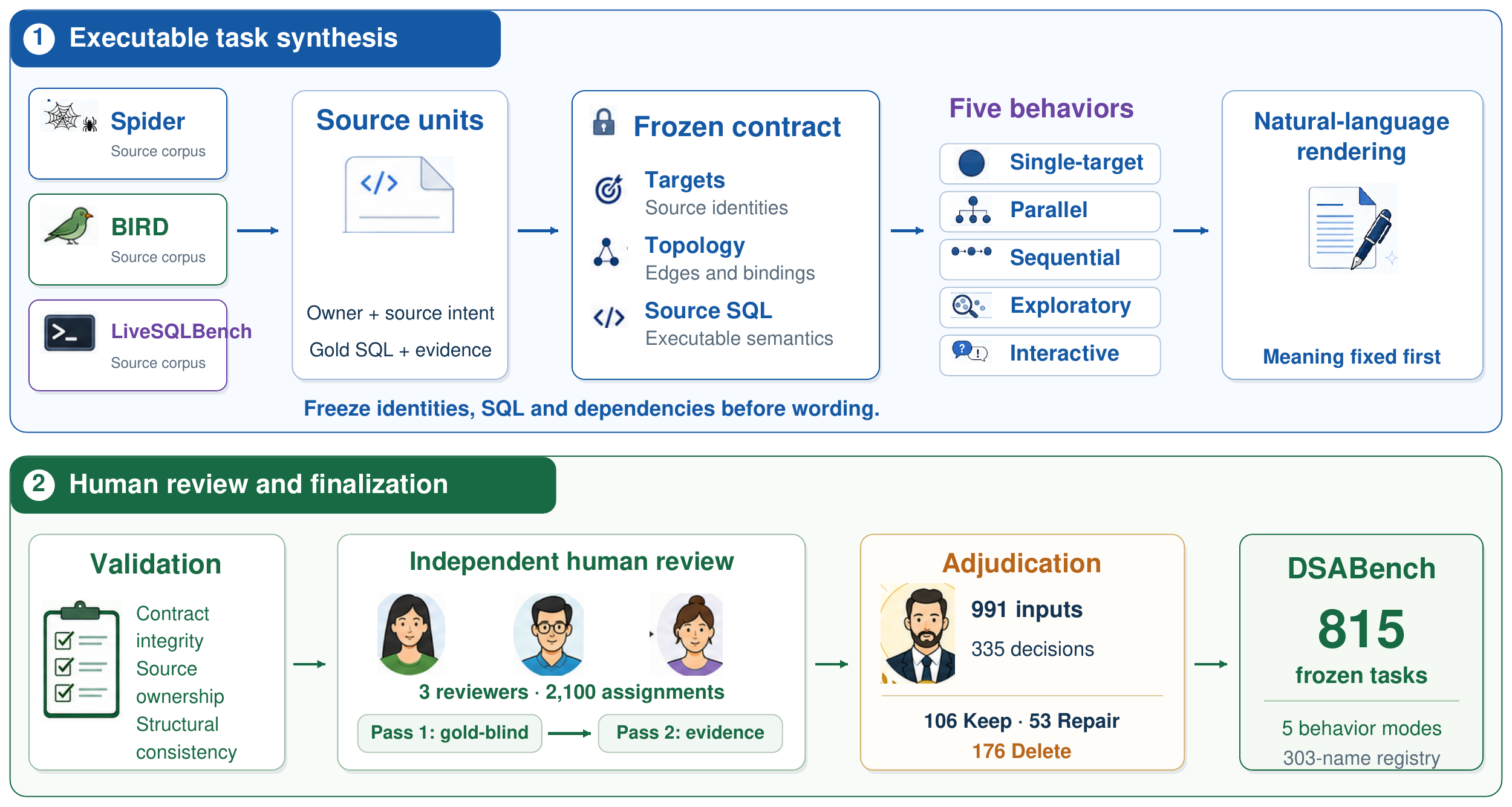}
\caption{Construction and human finalization. The two human passes are evidence stages within a review assignment, distinct from the number of independent reviewers. The 335 adjudicated cases are a subset of the 991-task finalization input; 176 deletions yield 815 tasks.}
\label{fig:construction}
\end{figure}

\paragraph{A concrete dependency.}
A Sequential task illustrates why this distinction matters. The user asks for vocal types in the song whose title matches a film with rental fee 0.99 and fewer than three inventory entries. The film database produces a title, which becomes an input to the song-database query. The gold SQL contains the literal \texttt{Badlands}, but the user never sees that value. Even after finding both databases, the agent must preserve the binding between the two queries. Appendix~\ref{app:example_sequential} presents the task and its executable binding; Figure~\ref{fig:workflow} shows the same kind of dependency schematically.

\paragraph{Human finalization.}
Three reviewers completed 2,100 assignments covering 1,356 candidate tasks: 744 candidates were reviewed independently twice and 612 once. Each assignment had two passes. In the first, reviewers saw only the request and public catalog evidence; they did not see gold sources, SQL, behavior labels, or contracts. After recording their first-pass judgment, they examined executable evidence in a second pass. Review and filtering left 991 candidates for finalization. An independent adjudicator then made 335 decisions: 106 to keep, 53 to repair only the question text, and 176 to delete. This produced the final 815 tasks (Figure~\ref{fig:construction}). Repairs were not allowed to change executable fields. Appendix~\ref{app:human_review} details the review coverage and available agreement evidence.

\paragraph{Evaluation boundary.}
The benchmark card documents composition, intended use, and limitations following datasheets and data statements~\citep{gebru2021datasheets,bender2018statements}.

\begin{table}[t]
\caption{Benchmark profile. All 815 tasks enter the main aggregate; the 710-task non-interactive slice is a sensitivity analysis. Registry size is distinct from final-task coverage.}
\label{tab:profile}
\centering\small
\begin{tabular}{llr}
\toprule
Dimension & Category & Tasks / value \\
\midrule
Corpus & Spider / BIRD / LiveSQLBench & 537 / 254 / 24 \\
Behavior & Single / Parallel / Sequential & 92 / 305 / 22 \\
 & Exploratory / Interactive & 291 / 105 \\
Multi-source & Jointly required (non-interactive) & 618 / 815 (75.8\%) \\
Target lists & More than one entry & 723 / 815 (88.7\%) \\
Difficulty & Easy / Medium / Hard & 65 / 516 / 234 \\
Frozen lists & Range / mean & 1--8 / 2.94 \\
Candidates & Range / mean & 11--166 / 66.08 \\
Registry & Names / scoped physical instances & 303 / 304 \\
\bottomrule
\end{tabular}
\end{table}

\section{Acquiring Evidence without Revealing the Targets}
A catalog-level task should let the agent inspect evidence rather than rely on database names. RouteDB stores summaries, schema descriptions, capabilities, entities, relationships, and provenance information. RouteEnv makes that information searchable through public database identifiers, while a private, dataset-scoped mapping resolves the corresponding execution targets. The annotated gold sources and SQL are never exposed to the router. All policies have access to the same types of public information, although they may use different prompts, tools, and stopping rules. We therefore treat the comparisons as reference conditions rather than equal-compute or equal-evidence experiments.

\paragraph{From evidence to a route.}
The router may search capabilities, inspect schemas, and examine relationships before returning its selected databases, source-local information requests, and coordination plan. A catalog relationship indicates that sources \emph{could} be used together, not that the user's task \emph{requires} them to be joined or sequenced. Inferring that requirement remains the router's job. This is the difference between navigating a catalog and constructing a task-specific workflow.

\paragraph{Relevant sources versus complete coverage.}
We first assess how much useful evidence retrieval can supply. In an audit of 20,776 source-unit records, Candidate Recall@50 is 99.56\%, reranked Recall@10 is 97.79\%, and MRR is 87.03\%. On the 815 formal tasks, Candidate Top-50 retrieves 93.83\% of annotated target-list entries on average, yet contains the entire target list for only 86.50\% of tasks; reranked Top-10 complete coverage falls to 57.18\%. These figures are from a report-only audit with no stated tuning on the 815 tasks; they do not establish held-out generalization or account for adaptive search. For Interactive tasks, retrieval is measured against the union of possible sources, unlike the branch-conditioned routing score. Whereas BEIR evaluates ranked retrieval across varied tasks~\citep{thakur2021beir}, our audit also measures whether the full annotated source set is retrieved (Appendix~\ref{app:retrieval_metrics}).

The difference matters especially for multi-source tasks. With reranked Top-10 results, Parallel tasks recover 83.77\% of required sources individually but all required sources for only 60.98\% of tasks. Exploratory tasks show a similar split: 76.58\% individual recall versus 44.67\% complete-set coverage. Strong source-level recall can therefore coexist with missing inputs that make the overall workflow impossible. This is why we report set coverage alongside conventional retrieval metrics.

\section{Evaluation Protocol and Experimental Design}
\label{sec:evaluation}
\paragraph{Headline metrics.}
We report four main metrics. \textbf{RA-Set} measures exact source-set recovery, $\mathbf1[P=G]$. \textbf{RA-Graph} also requires the normalized behavior/topology label and the applicable directed dependency edges. It is therefore a joint source-and-structure score, not a standalone measure of edge reasoning. \textbf{EA-Micro} measures correctness across required source contributions, counting missing outputs as incorrect. For branch $b$ of task $i$, let $t_{ib}$ indicate that all required contributions execute correctly (including verified sequential binding where applicable), $g_{ib}$ indicate RA-Graph correctness, and $h_{ib}$ be the sequential planning gate (or one for other tasks). Our end-to-end workflow measure is

\begin{equation}
\mathrm{Route{+}Exec\text{-}Graph}=\frac1N\sum_i\sum_{b\in B_i}w_{ib}g_{ib}t_{ib}h_{ib}
\end{equation}

Here $B_i$ contains one unit-weight episode for each ordinary task or two half-weight branches for an Interactive task (weighted by $w_{ib}$). We apply the gates before averaging the branches. This measure tests whether the annotated workflow succeeds; it does \emph{not} separately judge free-form answer synthesis or unrestricted conversation. We report precision and recall, directional routing errors, TEA/CEA, and set-conditioned metrics as supplementary diagnostics (Appendix~\ref{app:metrics}). Because the available aggregate results do not isolate label correctness from edge correctness, RA-Graph should not be interpreted as an edge-accuracy score.

\paragraph{Execution comparators.}
Execution correctness is evaluated with corpus-specific comparators. BIRD uses fixed source-local execution verdicts. For Spider, we compare ordered results when gold SQL contains \texttt{ORDER BY} and multisets otherwise. For LiveSQLBench-derived sources, we run against fixed PostgreSQL instances, normalize to two decimals using \texttt{ROUND\_HALF\_UP}, and compare unordered sets. These rules define correctness within DSABench; they are not claims of universal SQL equivalence or direct comparisons to official upstream benchmark scores. Consequently, pooled EA-Micro is a measure under this protocol, not a corpus-independent measure of SQL quality. Appendix~\ref{app:corpus_execution} details the comparators, checks, and corpus-level breakdown.

\paragraph{Routing study.}
We evaluate six routing policies across seven model backends on all 815 tasks. Random provides a chance-level reference. LLM Direct, Rerank, and Constrained LLM use fixed evidence schedules; ReAct and Hermes can inspect evidence iteratively. The policies share the same public information boundary but differ in how much evidence they see, how they use tools, how they are prompted, and when they stop. The results are therefore descriptive comparisons of reference policies, not compute-matched head-to-head tests. Full schedules and results are in Appendix~\ref{app:routing_prompts}. We also report a post hoc 710-task sensitivity analysis, which was not preregistered.

\paragraph{Factorial study.}
For the combined routing-and-execution study, we use four homogeneous backends (Claude, GPT, DeepSeek-V4-Flash, and Kimi) and vary five factors over $3\times2\times2\times2\times3$ configurations: routing policy (Direct/ReAct/Hermes), executor (Text-to-SQL/corpus-specific adapter), planner (Oracle/LLM), clarification (Never Ask/Ask Once), and local-request mode (privileged source/Global fused/Router local). Each configuration evaluates 710 ordinary tasks and 210 branch episodes, with branch weights returning the total to 815 base tasks. Our primary non-privileged comparisons fix LLM planning and summarize 12 configurations per backend and clarification policy. We report the privileged setting only as a diagnostic.

\paragraph{Clarification as an information intervention.}
In the Interactive protocol, the agent decides whether to ask a question before the hidden interpretation is revealed. We use that same decision for both latent branches. If the question matches the intended semantic slot, the simulator provides the branch-specific natural-language answer. The router can then revise its selected sources and local requests; the executor uses either the clarified request or a revised local request. The gold branch remains hidden throughout. Never Ask receives no answer but is tested against the same pair of possible targets. RA-Graph requires the \texttt{interactive} label, with no dependency edges, even when a branch has only one target database. This controlled simulator and its question-alignment judge limit what the experiment can establish (Appendix~\ref{app:source_recheck}).

Bootstrap intervals resample task clusters from recorded model outputs. They do not capture uncertainty from rerunning providers or random seeds, or every dependence caused by shared databases. The statistical unit and resampling assumptions therefore matter~\citep{dror2018significance}. We give evaluation-set-selected Best-of-grid results only as supplementary descriptive statistics: selecting a winner on the evaluation data can produce optimistic estimates~\citep{cawley2010selection}.

\flushbottom
\section{Results}
\subsection{Does Finding the Right Sources Suffice?}
\input{appendix/main_routing_table}
How far does correct source discovery take an agent? Table~\ref{tab:routing_main} gives routing results for all 815 tasks. The highest RA-Set score is 60.55\% (GPT/Hermes), while the highest RA-Graph score is 34.66\% (Claude/Hermes). These are maxima from different configurations in the routing-only study, scored before clarification. Every backend shows a noticeable drop when structure is required in addition to the source set. Results for Random and Rerank, along with the other reference policies, appear in Appendix~\ref{app:numeric}.

To see where workflow attempts break down, we examine one fixed non-oracle setting: Hermes routing, the corpus-specific adapter, LLM planning, Never Ask, and Router local requests. Across four backends and all 815 tasks, only 55.1\%--58.4\% of source-set-correct task weight also passes RA-Graph. Among graph-correct task weight, 32.3\%--37.0\% passes the full execution conjunction. Claude, for example, loses 200.5 task-equivalents when structural requirements are imposed and another 188 when execution is required (Table~\ref{tab:full_transitions}). Half-weight branches account for the fractional counts. These transitions locate failure stages but do not reveal the internal reasoning that caused them.

\begin{table}[h]
\centering\small
\caption{Expected task-equivalent success counts (half-weight branches) on one fixed non-oracle configuration over all 815 tasks: Hermes, corpus-specific adapter, LLM planner, Never Ask, and Router local. Routing-only and factorial evaluations are distinct experiments, so their routing counts can differ.}
\label{tab:full_transitions}
\begin{tabular}{lrrr}
\toprule
Backend & Set correct & Graph correct & Route+Exec correct \\
\midrule
Kimi K3 & 455 & 253 & 85 \\
GPT-5.6 Sol & 489.5 & 269.5 & 87 \\
Claude Opus 5 & 482.5 & 282 & 94 \\
DeepSeek-V4-Flash & 421.5 & 235 & 87 \\
\bottomrule
\end{tabular}
\end{table}

\subsection{Do Better Local Requests Yield Better Workflows?}
\begin{figure}[!t]
\centering
\includegraphics[width=\linewidth]{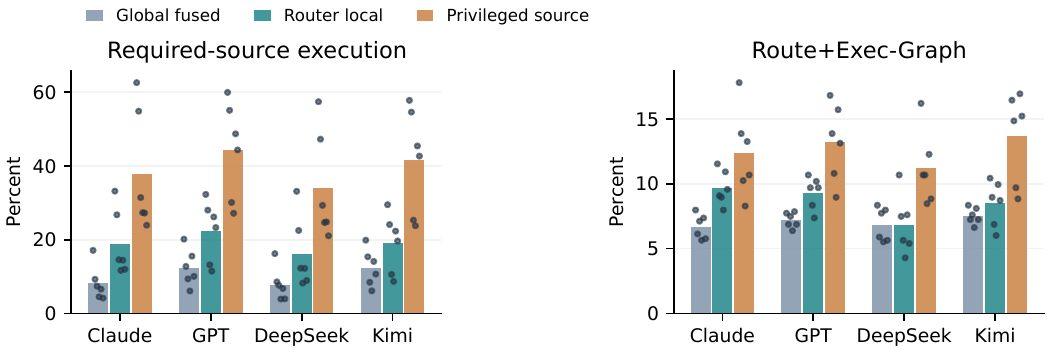}
\caption{Local request formulation improves source execution but does not ensure joint success. All 815 tasks; LLM planning and Never Ask fixed. Each bar averages the same six router/executor cells; dots are configuration observations, not confidence intervals. Privileged source questions supply additional evaluator-side information.}
\label{fig:querymode}
\end{figure}
Perhaps agents struggle because the global request does not make each database's role explicit. To test this, we compare passing the global request to every executor (Global fused) with passing each executor the source-local request produced by the router (Router local). We hold LLM planning and Never Ask fixed and average over the same six router/executor configurations. Router local yields 16.2\%--22.4\% mean EA-Micro, compared with 7.9\%--12.4\% for Global fused, improving source-level execution on every backend (Figure~\ref{fig:querymode}). Importantly, these local requests are generated by the agent from public inputs, not supplied by the evaluator.

That improvement at the source level does not reliably carry through to task completion. Mean Route+Exec-Graph is 6.9\%--9.7\% with Router local and 6.7\%--7.5\% with Global fused; for DeepSeek, the values are unchanged to two decimal places. EA-Micro credits correct individual contributions, but the end-to-end metric requires every contribution and the route structure to be correct at once. A source-level SQL score alone cannot tell us whether the task was completed.

We also evaluate privileged source questions as a diagnostic. With these evaluator-supplied inputs, EA-Micro reaches 34.1\%--44.2\% and Route+Exec-Graph reaches 11.2\%--13.7\%. Unlike Router local requests, these questions are not produced by the agent. Some include constraints that are absent from the visible user request (Appendix~\ref{app:examples}). The performance gap therefore reflects both possible request-decomposition failures and differences in what the inputs specify. It should not be read as the effect of improving the router alone.

\subsection{Can Clarification Repair Source Selection?}
Clarification addresses a different obstacle: the required database may depend on what the user meant. We compare Ask Once with Never Ask on the same 105 Interactive tasks, using LLM planning, non-privileged inputs, and the same 12 configurations per backend. Agents make the decision to ask before learning which interpretation is active, and both policies are scored on the same two possible source sets.

Asking one question increases exact branch-source recovery from 17.78\%--19.84\% to 58.25\%--61.43\% across backends, an improvement of 39.8--43.2 percentage points. The paired task-cluster intervals exclude zero (Appendix~\ref{app:interactive_corrected}). Because this comparison uses RA-Set, it is unaffected by the topology-label requirement. Among weighted observations in which the agent asks, 43.1\%--45.7\% change from an incorrect to a correct source set; only 1.6\%--2.5\% change in the opposite direction. The controlled reply often helps agents identify the right sources, although substantial errors remain.

On these tasks, Route+Exec-Graph increases from zero under Never Ask to 35.91\%--38.02\% under Ask Once. The zero baseline needs qualification: Never Ask obtains only 0--0.16\% RA-Graph, partly because the protocol requires an \texttt{interactive} label even when the active branch has a single target. Individual source executions do not all fail. We therefore regard exact source recovery as the clearest evidence of clarification's benefit, rather than attributing the entire end-to-end difference to better SQL generation or general dialogue ability.

\input{appendix/main_best_table}
\paragraph{Impact on the full benchmark.}
We also examine end-to-end performance over the full 815-task benchmark (Table~\ref{tab:best}). For each backend and clarification policy, the comparison covers 12 non-privileged configurations. The median is 6.69\%--7.98\% under Never Ask and 11.44\%--13.25\% under Ask Once; the largest observed values are 11.53\% and 16.99\%, respectively. Interquartile ranges describe variation across configurations, not uncertainty from different seeds or providers. The maxima are descriptive and have no correction for selection. Unlike the matched Interactive comparison, the full-benchmark contrast also includes ordinary tasks.

\paragraph{Sensitivity to task scope.}
The main finding is not driven solely by Interactive tasks. After excluding them, the best routing-only RA-Set and RA-Graph values rise to 66.20\% and 39.72\%; the highest non-privileged Never Ask conjunction reaches 13.24\%. The gap between source selection and structure, and the low full-workflow success rate, persist on the remaining 710 tasks (Appendix~\ref{app:noninteractive_sensitivity}). We treat the 22-task Sequential planning analysis as exploratory because of its small sample.

\section{Discussion and Limitations}
The two interventions tell a complementary story. Better local requests help agents execute individual contributions; clarification helps them choose the right sources when intent is ambiguous. Neither result, by itself, establishes that the entire workflow succeeds. Reporting both intermediate metrics and the joint score makes it possible to distinguish source, structure, and execution failures even when systems have similar end-to-end scores. It does not, however, reveal an agent's unobserved internal reasoning.

\begin{table}[!htbp]
\centering\small
\caption{Corpus-specific execution in the same fixed non-oracle configuration as Table~\ref{tab:full_transitions}. Ranges span four backends; corpus mix and execution comparators differ.}
\label{tab:corpus_main}
\begin{tabular}{lrrr}
\toprule
Corpus & Tasks & EA-Micro (\%) & Route+Exec-Graph (\%) \\
\midrule
Spider & 537 & 32.62--37.51 & 9.31--11.73 \\
BIRD & 254 & 22.78--26.07 & 11.81--14.17 \\
LiveSQLBench & 24 & 1.85--5.56 & 0.00--4.17 \\
\bottomrule
\end{tabular}
\end{table}

\paragraph{Why the pooled score needs context.}
Performance also varies with the source corpus (Table~\ref{tab:corpus_main}). In this fixed setting, Spider has higher EA-Micro than BIRD, while BIRD has a higher Route+Exec-Graph score. That reversal does not establish which corpus is intrinsically harder. The number of required sources, routing performance, and corpus-specific equality rules all affect the aggregate; these effects should not be mistaken for SQL-generation difficulty alone.

\paragraph{Scope.}
DSABench is an English-language research testbed built from public relational datasets, not a simulation of a full enterprise deployment. Its registry of 303 database names is smaller and more orderly than many production catalogs. The Sequential regime, which specifically tests value-flow dependencies, has only 22 tasks, and LiveSQLBench contributes 24. Public schemas may have appeared in model pretraining. Bootstrap intervals based on fixed outputs do not cover provider or seed variation or every dependence among tasks sharing a database. Finally, the routing policies are reference implementations with differing evidence and compute schedules, not faithful or budget-matched reproductions of earlier systems.

\paragraph{Language and protocol validity.}
Human review was intended to resolve unclear wording, but we did not measure its benefit against an unreviewed control group. Much of the Exploratory split retains standardized discovery phrasing, and several preserved examples have differences between the request and source-local requirements (Appendix~\ref{app:examples}). The available evidence does not support an all-task semantic-mismatch estimate or an inter-reviewer agreement statistic. Low execution scores may therefore reflect shortcomings in task wording as well as agent errors; we cannot separate these causes quantitatively. Interactive results also depend on two predefined interpretations and a semantic-slot alignment judge, rather than naturally occurring conversations or unrestricted answer generation.

\paragraph{Release and revision policy.}
The release retains the identities of all 815 tasks and specifies how branch-conditioned scores are computed. If future revisions change the visible requests, the affected tasks will need matched reevaluation; previously reported scores cannot simply be reinterpreted as improvements. Released task-level aggregates support numerical reanalysis. Exact reruns also depend on the original model outputs, service endpoints, databases, and software state.

\section{Conclusion}
DSABench evaluates a problem that begins before SQL is written: given a request and a catalog, can an agent identify the needed databases, coordinate their queries, and execute the complete workflow? The benchmark combines 815 reviewed tasks with public evidence tools and executable contracts to assess each stage and their conjunction. Our experiments show that source discovery is far from task completion. Better source-local requests do not always translate into a successful workflow, while one targeted clarification can substantially improve source recovery for ambiguous requests.

The value of the benchmark lies in making those distinctions visible. It measures where an agent makes progress and where that progress breaks down, while keeping the scope of its end-to-end score explicit: success on the annotated database workflow, not unrestricted final-answer correctness.

\flushbottom
\clearpage
\subsection*{AI use statement}
Generative AI tools were used to assist with coding and debugging, analysis scripts, interpretation and methodological review, literature search, drafting and language editing, translation, and figure preparation. Language models also helped render natural-language benchmark requests from executable contracts, as described in Section~\ref{sec:benchmark}. Generated requests underwent deterministic checks and human review; model output alone was never treated as ground truth. The authors take responsibility for the claims, citations, annotations, and results in this paper.

\subsection*{Ethics statement}
DSABench builds on existing research datasets, not private production databases. The licensing and redistribution terms of each upstream dataset still apply. Reviewer records are presented without personal identifiers. The benchmark does not establish that a tested agent is suitable for safety-critical or sensitive deployments, and source SQL is run under read-only controls.

\subsection*{Reproducibility statement}
Section~\ref{sec:evaluation} defines the metrics, experimental factors, and units of analysis. The appendix provides prompts, policy schedules, illustrative tasks, routing results, configuration summaries, and matched comparisons. The project release includes aggregate results and scripts to check the reported numerical summaries. Reproducing the exact model runs additionally requires their recorded outputs, software state, databases, generated SQL, and compatible model endpoints.
\input{refs.tex}
\clearpage
\appendix
\input{appendix/appendix_en}
\end{document}

%% file: math_commands.tex
\usepackage{amsmath,amsfonts,bm}

\def\eqref#1{equation~\ref{#1}}
\def\1{\bm{1}}

\DeclareMathAlphabet{\mathsfit}{\encodingdefault}{\sfdefault}{m}{sl}
\SetMathAlphabet{\mathsfit}{bold}{\encodingdefault}{\sfdefault}{bx}{n}

%% file: author_info.tex
\newcommand{\DSABenchPDFAuthors}{Jinlong Luo, Haoran Wu, Bo Xu}
\newcommand{\DSABenchAuthorBlock}{%
  Jinlong Luo \qquad Haoran Wu \qquad Bo Xu\thanks{Corresponding author.}\\
  Institute of Automation, Chinese Academy of Sciences\\
  Beijing, China%
}

%% file: appendix/main_routing_table.tex
\begin{table}[!htbp]
\centering
\caption{Routing results on all 815 frozen tasks. Methods use fixed method-specific evidence schedules and are therefore descriptive reference conditions, not compute-matched ablations.}
\label{tab:routing_main}
\begingroup
\definecolor{tbblue}{RGB}{231,241,252}
\definecolor{tbyellow}{RGB}{252,244,218}
\definecolor{tbgreen}{RGB}{230,244,235}
\definecolor{tbred}{RGB}{250,224,228}
\definecolor{tbstripe}{RGB}{247,247,247}
\fontsize{8.00}{9.15}\selectfont
\setlength{\tabcolsep}{1.40pt}
\renewcommand{\arraystretch}{1.24}
\begin{tabular}{@{}lrrrrrrrr@{}}
\toprule
\multirow{2}{*}{\textbf{Model Backend}} &
\multicolumn{2}{c}{\cellcolor{tbblue}\textbf{LLM Direct}} &
\multicolumn{2}{c}{\cellcolor{tbyellow}\textbf{Constrained LLM}} &
\multicolumn{2}{c}{\cellcolor{tbgreen}\textbf{ReAct}} &
\multicolumn{2}{c}{\cellcolor{tbred}\textbf{Hermes}} \\
\cmidrule(lr){2-3}\cmidrule(lr){4-5}\cmidrule(lr){6-7}\cmidrule(l){8-9}
& \cellcolor{tbblue!55}RA-Set $\uparrow$ & \cellcolor{tbblue!55}RA-Graph $\uparrow$
& \cellcolor{tbyellow!58}RA-Set $\uparrow$ & \cellcolor{tbyellow!58}RA-Graph $\uparrow$
& \cellcolor{tbgreen!58}RA-Set $\uparrow$ & \cellcolor{tbgreen!58}RA-Graph $\uparrow$
& \cellcolor{tbred!60}RA-Set $\uparrow$ & \cellcolor{tbred!60}RA-Graph $\uparrow$ \\
\midrule
Claude Opus 5 & 24.11 & 17.18 & 22.88 & 16.93 & 39.45 & 25.03 & 60.06 & 34.66 \\
\rowcolor{tbstripe}GPT-5.6 Sol & 23.74 & 16.32 & 24.17 & 16.93 & 49.82 & 30.67 & 60.55 & 33.01 \\
Kimi K3 & 21.78 & 15.83 & 20.25 & 14.85 & 52.15 & 30.31 & 55.83 & 30.92 \\
\rowcolor{tbstripe}DeepSeek-V4-Pro & 18.40 & 12.88 & 19.08 & 13.62 & 42.02 & 26.38 & 50.06 & 27.48 \\
DeepSeek-V4-Flash & 23.25 & 16.44 & 22.58 & 16.32 & 29.75 & 18.65 & 52.33 & 28.83 \\
\rowcolor{tbstripe}Qwen3.5-397B-A17B & 23.80 & 16.81 & 24.23 & 16.93 & 36.87 & 23.19 & 48.34 & 27.12 \\
Qwen3.6-35B-A3B & 18.04 & 13.25 & 15.83 & 12.02 & 37.18 & 23.44 & 47.48 & 26.87 \\
\midrule
Average (7 models) & 21.88 & 15.53 & 21.29 & 15.37 & 41.03 & 25.38 & 53.52 & 29.84 \\
\bottomrule
\end{tabular}
\par\vspace{1.5pt}
{\fontsize{7.05}{8.15}\selectfont\textbf{Notes:} Percent (\%), higher is better. RA-Set requires the exact required-database set. RA-Graph is a strict composite that additionally requires the normalized behavior/topology label and applicable directed dependency edges; it is not a pure dependency-edge metric. Interactive rows re-score the initial routing outputs against each branch with weight 0.5; no clarification is performed in this routing-only study.}
\endgroup
\end{table}

%% file: appendix/main_best_table.tex
\begin{table}[!htbp]
\centering\scriptsize
\caption{Non-privileged Route+Exec-Graph distributions on all 815 tasks (percent). LLM planning is fixed; each policy has 12 router/executor/fused-or-local cells. IQR describes configuration variation; maxima are descriptive.}
\label{tab:best}
\begin{tabular}{lrrrrr}\toprule
Backend & Never med. & Never IQR & Ask med. & Ask IQR & Ask max. \\
\midrule
Claude Opus 5 & 7.98 & 6.87--9.20 & 13.25 & 11.40--14.49 & 16.99 \\
GPT-5.6 Sol & 7.79 & 7.24--9.69 & 12.58 & 11.86--14.10 & 15.40 \\
DeepSeek-V4-Flash & 6.69 & 5.61--7.79 & 11.44 & 9.97--13.31 & 16.13 \\
Kimi K3 & 7.85 & 7.15--8.77 & 12.98 & 11.15--14.08 & 16.13 \\
\bottomrule\end{tabular}
\end{table}

%% file: refs.tex
% Shared English/Chinese bibliography; generated with the unmodified ICLR style.
\bibliographystyle{iclr2027_conference}
\bibliography{references}

%% file: appendix/appendix_en.tex
% DSABench Appendix -- public preprint version
% Intended placement: AFTER the bibliography in the public preprint.
% Public-preprint appendix.
%
% Suggested integration:
%   ... main paper ...
%   \input{statements_en}   % AI Use / Ethics / Reproducibility, if used by main paper
%   \input{refs}            % or \bibliography{...}
%   \appendix
%   \input{appendix/appendix_en}
%
% Packages assumed in the main preamble:
%   amsmath, amssymb, booktabs, fvextra

\section{Guide to the Supplementary Material}
\label{app:overview}

The appendix provides the task-construction, annotation, retrieval, metric, experimental, and reproducibility details behind the main results. All statistics refer to the same finalized benchmark and evaluation protocol used in the paper. We include the implementation and validation details needed to interpret the results without crowding the main presentation.

The benchmark contains 815 reviewed tasks drawn from a registry of 303 database names (304 dataset-scoped instances). All 815 contribute to the main results. Interactive tasks are scored over two equally weighted branches; we also report a post hoc analysis of the 710 non-interactive tasks. The corrected branch-conditioned results replace the earlier interaction evaluation. Recomputing these published aggregates did not require new model inference.

We distinguish the ability of the retrieval system to surface relevant sources from the agent's ability to select the exact set, recover the route structure, formulate local requests, and execute every required contribution. The combined workflow metric, \emph{Route+Exec-Graph}, tests whether these requirements hold together. It does not add a separate score for final natural-language answer generation.

\section{Benchmark Construction, Data, and Quality Control}
\label{app:construction}

\subsection{Benchmark Card}
\label{app:benchmark_card}

\subsubsection{Intended use and evaluation unit}

DSABench evaluates data agents that must work across multiple databases without being told which ones to use. Given a natural-language request, a candidate catalog, and RouteEnv's public metadata, an agent must identify the required databases and determine how to coordinate them. The benchmark measures exact source-set routing, the resulting route structure, the formulation of database-local requests, source-local Text-to-SQL execution, and the combined workflow outcome. It does not separately evaluate unconstrained final-answer generation.

The unit of evaluation is the full task, not an isolated schema, database, source unit, or SQL statement. DSABench is not designed to assess distributed transactions, production access control, schema drift, network federation, enterprise security, or high-stakes decision making.

\begin{table}[h]
\centering
\small
\caption{DSABench benchmark profile.}
\label{tab:app_benchmark_card}
\begin{tabular}{ll}
\toprule
Item & Frozen value \\
\midrule
Frozen reviewed records & 815 \\
Main quantitative scope & All 815 frozen tasks \\
Interactive base tasks (two branches) & 105 \\
Joint multi-source tasks & 618 / 815 (75.83\%) \\
Source corpora & Spider, BIRD, LiveSQLBench \\
Database registry & 303 names / 304 dataset-scoped instances \\
Final-task registry coverage & 263 names / 264 scoped instances \\
Target databases per task & 1--8; mean 2.9423 \\
Candidate databases per task & 11--166; mean 66.0798 \\
Task behaviors & Single, Parallel, Sequential, Exploratory, Interactive \\
Difficulty labels & Easy, Medium, Hard \\
Routing study & 7 model backends $\times$ 6 routing methods \\
Factorial evaluation grid & 4 model backends $\times$ 72 configurations \\
Primary routing metrics & RA-Set, RA-Graph \\
Primary conjunction metric & Route+Exec-Graph \\
\bottomrule
\end{tabular}
\end{table}

Each of the 105 Interactive tasks records two alternatives, not two databases that must be queried together. Accordingly, the 723/815 statistic for tasks with multi-entry target lists should not be interpreted as the prevalence of genuinely multi-source workflows. There are 618 tasks requiring multiple databases jointly: 75.83\% of the entire benchmark and 87.04\% of the 710 non-interactive tasks.

The full routing registry has 303 database names and 304 dataset-scoped instances. Of these, 263 names and 264 instances appear in at least one of the 815 final tasks. We report both figures to distinguish the space from which candidates are drawn from the subset directly represented in the evaluated tasks.

\subsubsection{Source-corpus composition}

\begin{table}[h]
\centering
\small
\caption{Source-corpus composition of the final 815-task set.}
\label{tab:app_source_composition}
\begin{tabular}{lrr}
\toprule
Source corpus & Final tasks & Observed database instances \\
\midrule
Spider & 537 & 166 \\
BIRD & 254 & 80 \\
LiveSQLBench & 24 & 18 \\
\midrule
Total & 815 & 264 \\
\bottomrule
\end{tabular}
\end{table}

We compose source units only within the same upstream corpus; a single DSABench task never mixes Spider, BIRD, and LiveSQLBench sources. This choice preserves the ownership and execution conventions of each corpus.

\subsubsection{Behavior and difficulty composition}

\begin{table}[h]
\centering
\small
\caption{Behavior and difficulty composition of the final benchmark.}
\label{tab:app_task_marginals}
\begin{tabular}{llr}
\toprule
Dimension & Category & Tasks \\
\midrule
Behavior & Single-target & 92 \\
         & Parallel & 305 \\
         & Sequential & 22 \\
         & Exploratory & 291 \\
         & Interactive & 105 \\
\midrule
Difficulty & Easy & 65 \\
           & Medium & 516 \\
           & Hard & 234 \\
\bottomrule
\end{tabular}
\end{table}

These counts describe the final benchmark rather than preset quotas. Generation, deterministic filtering, review, and adjudication can each change the distribution of task types and difficulty.

\subsubsection{Upstream data and contamination considerations}

Executable task material comes from Spider, BIRD, and LiveSQLBench. Distribution of the derived benchmark must respect each upstream resource's attribution and licensing terms; DSABench does not relicense the underlying databases. No private production databases are used.

Spider and BIRD are public datasets, and their schemas or source questions may have been encountered during model pretraining. We do not claim that every database is unseen. DSABench instead introduces a different evaluation setting: an agent must recover the complete set of participating sources, their applicable computational dependencies, and their executable contributions from the same public catalog evidence.

\subsection{Executable Source Units}
\label{app:source_units}

A source unit is the smallest construction element whose intended computation can be checked independently. We represent it as
\begin{equation}
  u = (d, q_u, s_u, E_u),
\end{equation}
where $d$ identifies the dataset-scoped owner database, $q_u$ describes the source-local intent in natural language, $s_u$ is read-only gold SQL, and $E_u$ includes schema and execution evidence for verification.

This representation ties each local intent to one physical database, keeps the SQL independently executable before composition, retains schema evidence for deterministic validation, and provides the reference local intent used in the privileged source-question diagnostic.

\subsection{Executable Task Contracts}
\label{app:frozen_contracts}

We fix the executable requirements of a task before generating its natural-language wording. For task $T$, the contract is
\begin{equation}
C_T = (D_T^\ast, U_T^\ast, \tau_T, G_T, B_T, A_T),
\end{equation}
where $D_T^\ast$ denotes the required database set, $U_T^\ast$ the required source units, $\tau_T$ the behavior category, $G_T$ the route and dependency structure, $B_T$ the relevant bindings or alternative branches, and $A_T$ the final aggregation or answer condition.

Once the contract is defined, its sources, queries, topology, dependency edges, sequential bindings, interactive branches, and intended computation remain fixed during language rendering. Human review then asks a separate question: does the resulting request describe those requirements clearly and faithfully? If an executable contract cannot be expressed naturally without changing its meaning, we regenerate or exclude the candidate instead of changing its answer retrospectively.

\subsection{Task Record and Information Partition}
\label{app:task_records}

Each finalized task has a self-contained evaluator record. Table~\ref{tab:app_task_record} groups its fields by who may see them. Physical database names appear here to explain the representation, but the router receives opaque \texttt{db\_*} identifiers. Keeping those identities and the answers private prevents the task record from acting as an oracle.

\begin{table}[h]
\centering
\footnotesize
\caption{Main fields and visibility of a DSABench task record.}
\label{tab:app_task_record}
\begin{tabular}{p{0.22\linewidth}p{0.25\linewidth}p{0.43\linewidth}}
\toprule
Partition & Representative fields & Purpose \\
\midrule
Public task input & user request, opaque candidate IDs & Model-visible request and candidate allowlist; task IDs remain bookkeeping metadata, not semantic evidence. \\
Frozen contract & source corpus, behavior, difficulty, target/source-unit references & Defines the task family and complete required source set. \\
Structural gold & topology, execution order, dependency edges, output-to-input bindings, clarification branches & Supports graph-, sequential-, and interaction-specific scoring. \\
Executable gold & source questions, read-only SQL, expected execution evidence, aggregation contract & Verifies each local contribution and complete task execution. \\
Review metadata & validation, review, adjudication, and repair decisions & Makes finalization decisions traceable without changing executable semantics. \\
Private identity & dataset-scoped physical map and evaluator-only annotations & Resolves public opaque IDs; never enters a router prompt. \\
\bottomrule
\end{tabular}
\end{table}

\subsection{Five Annotated Structural Regimes}
\label{app:behaviors}

\paragraph{Single-target.}
Exactly one database is required among distractors.  This isolates source
discovery from multi-source coordination.

\paragraph{Parallel.}
Multiple required databases can be queried independently before a comparison,
aggregation, or synthesis step.  No source output is required to instantiate
another source query.

\paragraph{Sequential.}
At least one verified output-to-input dependency is required: an upstream
result determines or instantiates a downstream source query.  A merely
preferred ordering is insufficient.

\paragraph{Exploratory.}
The task requires source discovery from incomplete or indirect evidence rather
than direct localization from a fully specified database-local intent.  The
frozen contract, rather than a lexical trigger, determines this label.

\paragraph{Interactive.}
The initial request contains genuine ambiguity such that different plausible
clarification outcomes map to different complete executable branches. The
initial wording must not already leak which branch is correct. These annotations define two branch-conditioned episodes per task in the corrected protocol (Appendix~\ref{app:interactive_corrected}). They enter the full-815 aggregate but not the non-interactive sensitivity slice.

\subsection{Exploratory Surface-Form Audit}
\label{app:surface_audit}

We also examined whether the wording itself gives away the type of task. Among the 291 Exploratory tasks, the exact phrase ``Identify the registered data sources needed to assess'' appears 99 times (34.0\%). A case-insensitive search for \texttt{registered data sources|source-grounded synthesis|identify.*sources} matches 242 tasks (83.2\%). This counts surface cues in the final requests; it is not a judgment that their underlying semantics are wrong. The repetition may nevertheless make the behavior category easier to infer. We therefore disclose it without silently changing the inputs on which our results were obtained. More varied paraphrases and wording-only repairs belong in a separately evaluated revision.

\subsection{Candidate Environments and Distractors}
\label{app:candidate_env}

Each task is presented with a catalog larger than its required source set. Candidate catalogs contain 11--166 databases (mean 66.0798), while a task requires one to eight (mean 2.9423). Distractors are designed to be semantically plausible under the same public metadata, rather than unrelated databases included only to increase catalog size. Candidate sets remain fixed across the reported evaluations.

\subsection{Deterministic Validation and Hard Gates}
\label{app:hard_gates}

We validate tasks both before and after generating their natural-language wording. Checks cover database ownership, schema references, read-only SQL, source necessity, contract consistency, behavior requirements, sequential dependencies, interactive branches, exploratory discovery semantics, and consistency among the final task and registry records. When a semantic check fails, the candidate is regenerated or removed; validation does not silently change the annotated gold route.

\subsection{Human Review and Independent Adjudication}
\label{app:human_review}

\subsubsection{Assignment design}

Task construction produced 1,356 candidates for human review. Three reviewers completed 2,100 assignments, 700 each. Of the candidates, 612 received one review and 744 received two independent reviews; none was assigned to all three reviewers. The available records do not preserve sufficient paired-review information to reconstruct agreement reliably or to determine the reviewer-coverage distribution after final filtering. We therefore report the observed coverage without presenting it as a reliability estimate.

\subsubsection{Pass 1: gold-blind review}

In the first pass, reviewers see an anonymous item identifier, the user request, the candidate databases, and the same schema and capability evidence available to the router. They do not see the source corpus, behavior label, gold sources, source questions, gold SQL, execution evidence, or task contract. They judge whether the request is natural and complete, which sources seem necessary, whether the target set appears unique, and which behavior it suggests. They also record confidence and free-form notes. Their judgments are saved before the second pass begins.

\subsubsection{Pass 2: evidence-based verification}

In the second pass, reviewers can inspect the required sources, source questions, gold SQL, dependencies and execution order, and any composition, exploration, or ambiguity constraints. They check the task as a whole and each source's contribution. Successful SQL execution is not sufficient if the request, required sources, queries, coordination structure, and intended composition disagree.

\begin{table}[h]
\centering
\footnotesize
\caption{Human-review and adjudication fields used during finalization.}
\label{tab:app_review_fields}
\begin{tabular}{p{0.16\linewidth}p{0.36\linewidth}p{0.38\linewidth}}
\toprule
Stage & Required judgments & Evidence boundary \\
\midrule
Pass 1 & naturalness/completeness; minimum sufficient source set; apparent uniqueness and behavior; confidence; notes & User request and router-visible catalog evidence; gold source set, SQL, behavior, and contract hidden. \\
Pass 2 & gold-set validity/uniqueness; behavior and multi-source logic; per-source question preservation; SQL/DB mapping; overall validity and severity & Frozen target set, source questions, SQL, dependencies/order, and behavior-specific contract revealed after Pass 1 lock. \\
Adjudication & \textsc{Keep}, \textsc{Repair}, \textsc{Delete}, or expert escalation; severity and rationale & Full audit trail; \textsc{Repair} is restricted to surface wording while every executable field remains frozen. \\
\bottomrule
\end{tabular}
\end{table}

\subsubsection{Adjudication and finalization}

After the initial reviews and deterministic filtering, 991 candidate tasks remained. An independent adjudicator examined selected risks, defects, disagreements, and quality-control cases, making 335 task-level decisions.

\begin{table}[h]
\centering
\small
\caption{Independent adjudication outcomes.}
\label{tab:app_adjudication}
\begin{tabular}{lr}
\toprule
Decision & Tasks \\
\midrule
KEEP & 106 \\
REPAIR & 53 \\
DELETE & 176 \\
\midrule
Total adjudicated & 335 \\
Final benchmark after finalization & 815 \\
\bottomrule
\end{tabular}
\end{table}

A repair changes only the user-facing question. It cannot alter the required databases, local questions, SQL, dependencies, execution order, topology, or other executable requirements. If a correction would change the task's semantics, the candidate is excluded rather than converted into a different task.

\section{RouteDB, RouteEnv, and Retrieval Coverage}
\label{app:routeenv}

\subsection{Benchmark-Side Routing Environment}

RouteDB stores benchmark-side routing knowledge about registered databases,
including schema and column semantics, capabilities, source relationships,
semantic cards, provenance, and stable public identifiers.  RouteEnv exposes
this knowledge through a controlled tool interface.

RouteEnv standardizes the observable information and action boundary, not the
internal routing trajectory.  Different routers may search, inspect, gather
evidence, maintain hypotheses, revise those hypotheses, and stop at different
times while remaining comparable against the same benchmark substrate.

\subsection{Gold-Hidden Information Boundary}

Evaluated routers do not receive gold target database sets, source-unit gold
SQL, final answers, hidden behavior labels used by scoring, evaluator-private
physical identity maps, or evaluator-only scoring annotations.  Model-visible
database identities are stable public identifiers; physical identities are
resolved by an evaluator-private dataset-scoped map, avoiding ambiguity when a
bare database name exists in more than one source dataset.

\subsection{Retrieval Configuration}

The retrieval system uses hierarchical IDF with semantic rescue, MiniLM semantic embeddings, and a fixed reranker.  The intrinsic retrieval audit uses a candidate pool of 50 and a reranked shortlist of 10. Runtime policies can issue additional calls and inspect other allowed candidates; these are not universal episode-level source caps.  The 815 formal tasks
are report-only for retrieval evaluation and are not used to tune retrieval or
reranking.

\subsection{Retrieval-Substrate Metrics}
\label{app:retrieval_metrics}

For task $i$, let $G_i$ denote the complete gold database set and $C_i^K$ the
retrieved Top-$K$ candidate set.  We report
\begin{equation}
\mathrm{TargetRecall@K}
=
\frac{1}{N}\sum_{i=1}^{N}
\frac{|G_i\cap C_i^K|}{|G_i|},
\label{eq:target_recall}
\end{equation}
which measures the average fraction of required sources retrieved, and the
stricter
\begin{equation}
\mathrm{AllTargetCoverage@K}
=
\frac{1}{N}\sum_{i=1}^{N}
\mathbf{1}[G_i\subseteq C_i^K],
\label{eq:all_target_coverage}
\end{equation}
which requires every database needed to solve the task to appear in the Top-$K$
set.  These are retrieval-substrate metrics and are intentionally separate from
agent routing metrics such as RA-Set and RA-Graph.

\subsection{RouteEnv Retrieval Results}
\label{app:routeenv_results}

\subsubsection{Single-source audit}

On 20,776 source-unit audit records, Candidate Recall@50 is 99.56\%,
reranked Recall@10 is 97.79\%, and MRR is 87.03\%.  These single-source values
do not imply complete multi-source readiness because one missing required source prevents completeness if a policy commits using only that retrieved set. Adaptive searches can expose additional evidence.

\subsubsection{Formal-task multi-target coverage}

\begin{table}[h]
\centering
\small
\caption{Frozen RouteEnv retrieval results on all 815 formal tasks.}
\label{tab:app_routeenv_overall}
\begin{tabular}{lrrr}
\toprule
Stage & $K$ & TargetRecall@K & AllTargetCoverage@K \\
\midrule
Candidate & 10 & 75.04\% & 51.53\% \\
Candidate & 20 & 85.09\% & 66.99\% \\
Candidate & 50 & 93.83\% & 86.50\% \\
Reranked  & 3  & 55.66\% & 30.43\% \\
Reranked  & 5  & 66.76\% & 41.84\% \\
Reranked  & 10 & 78.63\% & 57.18\% \\
\bottomrule
\end{tabular}
\end{table}

The gap between TargetRecall@K and AllTargetCoverage@K is the relevant signal
for multi-database routing: a retriever can recover most individual required
databases while still failing to provide a complete source set for many tasks.

\subsubsection{Breakdown by source corpus}

\begin{table}[h]
\centering
\small
\caption{Formal-task retrieval coverage by source corpus.}
\label{tab:app_routeenv_dataset}
\begin{tabular}{lrrrr}
\toprule
& \multicolumn{2}{c}{Candidate Top-50} & \multicolumn{2}{c}{Reranked Top-10} \\
Dataset & TargetRecall & AllCoverage & TargetRecall & AllCoverage \\
\midrule
BIRD & 100.00\% & 100.00\% & 91.21\% & 81.89\% \\
LiveSQLBench & 100.00\% & 100.00\% & 91.32\% & 79.17\% \\
Spider & 90.64\% & 79.52\% & 72.11\% & 44.51\% \\
\bottomrule
\end{tabular}
\end{table}

The largest corpus-specific retrieval deficit is concentrated in Spider rather
than being uniform across all three corpora.

\subsubsection{Breakdown by benchmark behavior}

\begin{table}[h]
\centering
\small
\caption{Formal-task retrieval coverage by behavior.}
\label{tab:app_routeenv_behavior}
\begin{tabular}{lrrrr}
\toprule
& \multicolumn{2}{c}{Candidate Top-50} & \multicolumn{2}{c}{Reranked Top-10} \\
Behavior & TargetRecall & AllCoverage & TargetRecall & AllCoverage \\
\midrule
Single-target & 98.91\% & 98.91\% & 98.91\% & 98.91\% \\
Parallel & 97.14\% & 90.82\% & 83.77\% & 60.98\% \\
Sequential & 100.00\% & 100.00\% & 95.45\% & 90.91\% \\
Exploratory & 96.37\% & 84.19\% & 76.58\% & 44.67\% \\
Interactive & 71.43\% & 66.67\% & 48.10\% & 37.14\% \\
\bottomrule
\end{tabular}
\end{table}

Interactive tasks are the largest behavior-specific retrieval bottleneck.  This
is consistent with their design: relevant sources cannot always be localized
from one fully specified intent before ambiguity resolution.

\subsubsection{Breakdown by gold target-set size}

\begin{table}[h]
\centering
\small
\caption{Formal-task retrieval coverage by target-set size.}
\label{tab:app_routeenv_goldsize}
\begin{tabular}{lrrrr}
\toprule
& \multicolumn{2}{c}{Candidate Top-50} & \multicolumn{2}{c}{Reranked Top-10} \\
$|G|$ & TargetRecall & AllCoverage & TargetRecall & AllCoverage \\
\midrule
1 & 98.91\% & 98.91\% & 98.91\% & 98.91\% \\
2--3 & 92.09\% & 88.75\% & 77.13\% & 62.99\% \\
4--5 & 96.75\% & 86.62\% & 79.84\% & 40.76\% \\
6--10 & 92.84\% & 57.14\% & 61.45\% & 3.90\% \\
\bottomrule
\end{tabular}
\end{table}

Complete-set coverage deteriorates much faster than partial target recall as the
number of required sources grows.  Single-target Recall@K therefore materially
overstates readiness for multi-source tasks.

\subsection{Routing Output Contract}
\label{app:routing_output}

A DSABench routing prediction is more than a relevance ranking.  Depending on
the behavior, a normalized route may include a selected database set, route
topology, sequential dependency edges and execution order, interactive branch
resolution, and database-specific subqueries/information requests.  The
benchmark thus evaluates executable workflow induction rather than schema
localization alone.

\section{Evaluation Metrics}
\label{app:metrics}

\subsection{Route-Set Selection}

For predicted database set $P$ and gold set $G$, exact route-set accuracy is
\begin{equation}
\mathrm{RA\text{-}Set}=\mathbf{1}[P=G].
\end{equation}
We additionally report database-level precision, recall, and F1.  Directional
selection errors are
\begin{equation}
\mathrm{OR}=\frac{|P\setminus G|}{|P|},
\qquad
\mathrm{UR}=\frac{|G\setminus P|}{|G|},
\end{equation}
where OR measures unnecessary source selection and UR measures missing required
sources.  For an empty prediction, precision, F1, and OR are zero, recall is zero, and UR is one. These rows remain in all task-level averages. Consequently OR is not always one minus macro precision.

\subsection{Graph-Level Routing}

RA-Graph requires RA-Set correctness and all applicable structural constraints,
specifically the normalized behavior/topology label and canonical directed dependency-edge set. The label itself is part of this score; evidence quality and a final natural-language answer are not independently checked by RA-Graph.
It is intentionally stricter than RA-Set.

\subsection{Source Execution}

Let $U_T^\ast$ denote the required source units for task $T$.  EA-Micro is the
micro-average fraction of required source-unit executions that match gold.
Task Execution Accuracy (TEA) requires every required source unit to be correct and valid bindings for dynamic sequential execution.  Execution coverage is reported separately when needed so technical
failure to produce executable SQL is not conflated with a semantically wrong
execution result.

\subsection{Conditional Execution}

Conditional execution describes execution on the subset with correct routing;
this subset may differ in difficulty across systems:
\begin{equation}
\mathrm{CEA\text{-}Set}
=\mathbb{E}[\mathrm{TEA}\mid\mathrm{RA\text{-}Set}=1],
\end{equation}
\begin{equation}
\mathrm{CEA\text{-}Graph}
=\mathbb{E}[\mathrm{TEA}\mid\mathrm{RA\text{-}Graph}=1].
\end{equation}

\subsection{Route+Execution Conjunction Metrics}

For task $i$, let $s_i$ and $g_i$ be set and graph correctness, $t_i$ be TEA, and $h_i$ be the sequential plan gate (one for other behaviors). The canonical evaluator computes $s_it_ih_i$ for the set conjunction and $g_it_ih_i$ for the graph conjunction. We report the graph conjunction as \emph{Route+Exec-Graph}. For Interactive, these gates use the latent branch target and gold, and their conjunction is averaged across branches. There is no independent free-form final-answer or general dialogue-quality gate.

\subsection{Execution Comparators}
The 288-configuration table is obtained by read-only rescoring of the fixed inference outputs, with no new model calls. BIRD uses fixed source-local execution verdicts. Spider uses ordered equality when the gold SQL contains \texttt{ORDER BY}; otherwise duplicate multiplicities are preserved with multiset (Counter) equality. LiveSQLBench-derived sources use fixed PostgreSQL instances, numeric normalization to two decimals with \texttt{ROUND\_HALF\_UP}, and unordered set equality. Set equality discards duplicate multiplicity for that corpus; these results are protocol-specific rather than interchangeable with Spider multiset scores or official upstream leaderboards.

Missing SQL, parser failures, and execution errors score zero. Dynamic sequential rows use validated bound SQL. Validation covers 264,960 episode-level records and 234,720 base-task rows across 288 configurations. Publication rescoring makes no new model calls, does not alter the evaluated task definitions, and passes the four reported hard-error checks. Technical SQL failures remain nonzero; those four checks do not certify all runtime behavior or natural-language validity.

Each cell includes 815 base tasks and 2,293 weighted required source contributions. Main question-mode comparisons fix LLM planning and Never Ask and average six router/executor cells. Because denominators match across cells, averaging cell EA-Micro equals pooling their correct-source counts and denominators. It is not the mean of task-normalized execution ratios. An independent check recomputes 2,304 aggregate metric values from the new task-level table, with maximum absolute discrepancy below $10^{-15}$. The manuscript update does not re-execute SQL verdicts.

\subsection{Corpus-Specific Execution Diagnostic}
\label{app:corpus_execution}
Because the canonical execution comparator is corpus-specific, pooled EA-Micro is
best read as a DSABench-protocol aggregate rather than a universal SQL-equivalence
score. Table~\ref{tab:corpus_execution_diag} makes this heterogeneity visible for
one fixed non-oracle reference slice (Hermes + corpus-specific adapter + LLM
planner + Never Ask + Router local) using the fixed full-815 aggregates.
The rows are not intended as cross-corpus difficulty rankings: task mix and
comparator semantics differ, and the 24-task LiveSQLBench slice is especially
small.

\begin{table}[h]
\centering
\small
\caption{Corpus-specific execution diagnostics for one fixed non-oracle evaluation slice. Ranges are across the four factorial backends. These full-815 aggregates expose corpus heterogeneity under the main reporting scope.}
\label{tab:corpus_execution_diag}
\setlength{\tabcolsep}{4.5pt}
\begin{tabular}{lrrrr}
\toprule
Corpus & Tasks & EA-Micro & TEA & Route+Exec-Graph \\
\midrule
Spider & 537 & 32.62--37.51\% & 14.71--18.81\% & 9.31--11.73\% \\
BIRD & 254 & 22.78--26.07\% & 12.80--15.35\% & 11.81--14.17\% \\
LiveSQLBench & 24 & 1.85--5.56\% & 0.00--4.17\% & 0.00--4.17\% \\
\bottomrule
\end{tabular}
\end{table}

These values are recomputed from the task-level evaluation outputs with
zero new model calls or SQL executions. They show why a pooled execution number
should not be interpreted as corpus-invariant capability; no causal explanation
for the cross-corpus differences is claimed.

\subsection{Sequential and Interactive Diagnostics}

Sequential tasks additionally report plan accuracy.  Planner comparisons are
conditioned on route-set correctness so a missing database is not counted again
as a planning error.  Interactive reporting includes Ask rate, clarification-
handler success, effective route change given Ask, exact route-set repair given
Ask, and exact route-set harm given Ask.  These mechanism metrics distinguish
asking a clarification question from using its answer to revise the source
hypothesis.

\section{Experimental Protocols}
\label{app:protocols}

\subsection{Seven-Backend Routing Study}
\label{app:routing_study}

The routing-only study evaluates every combination of seven formal model
backends and six routing methods on all 815 tasks:
\begin{equation}
7\times 6\times 815=34{,}230
\end{equation}
task-level observations.

\paragraph{Model backends.}
Kimi K3, DeepSeek-V4-Pro, GPT-5.6 Sol, and Claude Opus 5, together with
Qwen3.5-397B-A17B, DeepSeek-V4-Flash, and Qwen3.6-35B-A3B.

\begin{table}[h]
\centering
\footnotesize
\caption{Frozen routing-study model and runtime controls. Context limits are
the configured endpoint limits, not guaranteed prompt lengths.}
\label{tab:app_model_controls}
\begin{tabular}{lrr}
\toprule
Backend & Configured context & Sampling temperature \\
\midrule
Kimi K3 & 1,048,576 & 1.0 (provider override) \\
DeepSeek-V4-Pro & 1,048,576 & 0.0 \\
GPT-5.6 Sol & 262,144 & 0.0 \\
Claude Opus 5 & 262,144 & 0.0 \\
Qwen3.5-397B-A17B & 262,144 & 0.0 \\
DeepSeek-V4-Flash & 262,144 & 0.0 \\
Qwen3.6-35B-A3B & 131,072 & 0.0 \\
\bottomrule
\end{tabular}
\end{table}

The inspected routing defaults specify four task workers for remote endpoints and eight for local endpoints, with Hermes episode defaults of two and four respectively. These are configuration defaults, not independently verified realized concurrency in every experiment.
ReAct and Hermes both permit at most 30 environment turns.  Parser, provider,
or tool failures remain scored failures, and successful earlier termination is
permitted only after a valid final route is parsed.

\paragraph{Routing methods.}
Random, Rerank, LLM Direct, Constrained LLM, ReAct, and Hermes.

No downstream Data Agent participates in the routing-only score.  All methods
operate against equivalent frozen RouteEnv content.  Technical/provider/parser
tool failures remain in the denominator rather than being silently discarded.

\subsection{Routing Baseline Semantics}
\label{app:routing_semantics}

\paragraph{Random.}
A chance-level source-selection reference over the available candidate environment.

\paragraph{Rerank.}
Uses a fixed RouteEnv retrieval/inspection schedule followed by an LLM route decision. This is a retrieval-conditioned LLM baseline, not a deterministic or LLM-free retrieval-only baseline.

\paragraph{LLM Direct.}
A one-shot semantic routing decision.

\paragraph{Constrained LLM.}
An LLM route using a fixed candidate-inspection schedule and the same structured final decision contract. The inspected code visits candidates in allowlist order up to the available inspection budget; it does not guarantee inspection of the entire catalog.

\paragraph{ReAct.}
A generic iterative reason--act agent operating against the RouteEnv tool
boundary.

\paragraph{Hermes.}
A routing-specific tool-using episode that can search, inspect, maintain a route
hypothesis, revise it, and submit a structured route.  Hermes is an evaluated
reference system rather than part of the benchmark definition.

\subsection{Homogeneous 72-Cell Route+Execution Study}
\label{app:factorial}

The homogeneous study uses Claude Opus 5, GPT-5.6 Sol, DeepSeek-V4-Flash, and
Kimi K3.  Each model is evaluated over
\begin{equation}
3\ \text{routers}
\times 2\ \text{data agents}
\times 2\ \text{planners}
\times 2\ \text{interaction policies}
\times 3\ \text{query modes}
=72
\end{equation}
configurations, yielding
\begin{equation}
4\times72\times815=234{,}720
\end{equation}
task-level observations.

\begin{table}[h]
\centering
\small
\caption{The homogeneous 72-cell factorial design.}
\label{tab:app_72_design}
\begin{tabular}{lll}
\toprule
Factor & Values & Interpretation \\
\midrule
Router & LLM Direct / ReAct / Hermes & source discovery and route induction \\
Data Agent & Text-to-SQL / corpus-specific adapter & downstream source-level SQL \\
Planner & Oracle / LLM & sequential source ordering \\
Interaction & Never Ask / Ask Once & interactive clarification policy \\
Query mode & Privileged / Fused / Router & database-local request representation \\
\bottomrule
\end{tabular}
\end{table}

\paragraph{Privileged source.}
The downstream executor receives the frozen source-unit question for a selected
database.  This is a privileged-input execution diagnostic rather than a deployable routing condition or a mathematical upper bound on all possible systems.

\paragraph{Global fused.}
The visible global request is sent to every selected database without local decomposition; after valid clarification it includes the visible clarification exchange.

\paragraph{Router subquery.}
The router-generated database-specific request is used downstream.

\subsection{Matched Comparisons and Statistical Protocol}
\label{app:statistics}

Factor effects are estimated using matched comparisons that hold every
non-target factorial dimension fixed.  Supplementary marginal query-mode effects use all tasks; the main question-mode plot instead fixes LLM planning and Never Ask.
Planner effects are interpreted on Sequential tasks and conditioned on route-set
correctness.  Interaction effects are interpreted on Interactive tasks and are
additionally stratified by router.

Confidence intervals used by matched-effect figures are generated from 5,000
task-level cluster-bootstrap resamples.  The benchmark task is resampled as a cluster, retaining matched configurations and source contributions. This captures task-sampling uncertainty conditional on fixed experiment outputs; it does not estimate independent rerun/provider variation or all correlations induced by shared databases.

Best-of-72 selects the highest Route+Exec-Graph configuration independently within each
model's own 72-cell space.  It is a selected descriptive maximum, not a held-out fixed-cell comparison. Supplied winner intervals are conditional on the selected cell and are not selection-corrected; we omit them from the manuscript winner table. Planner and interaction variants can tie, so a chosen variant need not be uniquely optimal.

\section{Prompts, Tools, and Execution Policies}
\label{app:prompts}

This section records the fixed instruction text and machine interfaces used for
construction and evaluation.  Task-specific catalog observations, candidate
lists, schemas, and user requests are denoted by angle-bracket placeholders;
they are serialized deterministically at runtime.  The templates were audited
for consistency across all reported experiments.

\subsection{Contract-to-Language Renderer}
\label{app:renderer_prompt}

The construction model is instructed to render the precomputed contract rather than create new executable annotations.  Its system message is
``Return exactly one valid JSON object. Do not use markdown fences or add
commentary.''  The user template is:

\begin{Verbatim}[fontsize=\scriptsize,breaklines=true,breakanywhere=true]
You are rendering a verified DSABench task plan. Learn the transformation
pattern from the one-shot example, but do not copy its domain, wording,
entities or schema.

ONE-SHOT EXAMPLE INPUT:
<behavior-matched synthetic verified plan>
ONE-SHOT EXAMPLE OUTPUT:
<behavior-matched synthetic JSON rendering>

TARGET VERIFIED PLAN:
<frozen plan, source units, topology, bindings, branches, aggregation>

MANDATORY LITERALS:
<locked entities, numbers, years, operations, and output fields>

HARD RULES:
- Treat the verified plan, schema slices, knowledge, edges, and aggregation
  as the specification.
- Produce one new user intent requiring the specified topology; do not
  concatenate, quote, enumerate, or lightly paraphrase source questions.
- Do not expose database/source IDs, SQL, or table/column names unless they
  are genuinely public concepts in the user request.
- Preserve every locked literal and operation; invent no fact or constraint.
- Make every source indispensable and cover it in the contribution map.
- Use external knowledge only when the verified plan requires it.
- Return JSON only and follow the requested output schema.

OUTPUT SCHEMA:
<behavior-specific JSON schema>

<optional previous hard-check feedback>
\end{Verbatim}

Behavior-specific hard rules further require: (i) Parallel sources to execute
independently before one derived synthesis, rather than a side-by-side list;
(ii) Sequential wording to express one declarative outcome without revealing
the execution procedure, while removal of the upstream result makes the
downstream request uninstantiable; (iii) Exploratory wording to require
capability discovery without target IDs or unique schema literals; and
(iv) Interactive wording to encode a genuine ambiguity whose natural response
changes only the disputed slot.  The one-shot examples bundled with the
renderer are synthetic and behavior matched; they are distinct from the frozen
benchmark examples in Section~\ref{app:examples}.

\subsection{RouteEnv Action Interface}
\label{app:routeenv_actions}

All adaptive routers act through the signatures in
Table~\ref{tab:app_routeenv_actions}.  Each call returns observations with
stable evidence IDs; the environment exposes facts but never selects a route.

\begin{table}[h]
\centering
\scriptsize
\caption{Public RouteEnv actions. Optional arguments are shown in brackets.}
\label{tab:app_routeenv_actions}
\begin{tabular}{p{0.35\linewidth}p{0.55\linewidth}}
\toprule
Action & Function \\
\midrule
\nolinkurl{search_catalog(query[, dataset, top_k])} & Retrieve database candidates from the public catalog. \\
\nolinkurl{inspect_database(database_id[, table_limit])} & Read a database summary, capabilities, and table overview. \\
\nolinkurl{inspect_table(table_id[, column_limit])} & Inspect a public table card and columns. \\
\nolinkurl{inspect_column(column_id)} & Inspect one public column card. \\
\nolinkurl{search_capability(query[, dataset, top_k])} & Search normalized capabilities. \\
\nolinkurl{search_entity(query[, dataset, top_k])} & Search catalog entities. \\
\nolinkurl{check_coverage(database_id)} & Summarize evidence coverage for a database. \\
\nolinkurl{expand_relations(object_type, object_id[, relation_types, depth, limit])} & Traverse public catalog relationships. \\
\nolinkurl{check_compatibility(source_database_id, target_database_id[, limit])} & Inspect possible cross-source compatibility evidence. \\
\bottomrule
\end{tabular}
\end{table}

The router submits a structured route plan. The validator requires the following
shape; database and evidence IDs must come from the candidate allowlist and
observations, respectively.

\begin{Verbatim}[fontsize=\scriptsize,breaklines=true,breakanywhere=true]
{
  "selected_databases": ["db_..."],
  "execution_plan": {
    "topology": "single|parallel|sequential|exploratory|interactive",
    "nodes": [
      {"database_id": "db_...", "information_request": "..."}
    ],
    "edges": [
      {"from": "db_...", "to": "db_...", "type": "...",
       "reason": "..."}
    ]
  },
  "evidence_ids": ["ev_..."],
  "confidence": 0.0
}
\end{Verbatim}

\subsection{Routing Baseline Prompts and Schedules}
\label{app:routing_prompts}

\paragraph{Fixed-evidence LLM routes.}
LLM Direct, Rerank, and Constrained LLM differ in evidence acquisition but use
the same final decision instruction.  Direct inspects up to three retrieved databases and Rerank up to ten, subject to the turn budget. Constrained LLM inspects candidates in allowlist order up to the budget (\texttt{max\_steps - 2} in the inspected implementation). All three then make this LLM decision call. Search results are filtered to the candidate allowlist.

\begin{Verbatim}[fontsize=\scriptsize,breaklines=true,breakanywhere=true]
Select necessary databases using only RouteEnv observations.
Candidate allowlist: <db IDs>
User query: <query>
Observations: <evidence-bearing observations>
Valid evidence IDs: <ev IDs>

Return only route plan JSON fields. A single route uses one database; a
parallel route uses independent sources; a sequential route requires an
output-to-input dependency; an exploratory route requires source discovery;
an interactive route represents the clarified executable branch.
\end{Verbatim}

\paragraph{ReAct.}
The generic agent receives at most 30 turns.  Its fixed system instruction is:

\begin{Verbatim}[fontsize=\scriptsize,breaklines=true,breakanywhere=true]
You are a database routing model operating in DSABench RouteEnv.
Database IDs are opaque. Discover facts through one action at a time. The
environment provides facts but never decides the route. Do not infer an
execution order merely because two columns are compatible. Cite direct
evidence for every selected database.

Emit exactly one block per turn:
<ACTION>...</ACTION>
<ASK_USER>...</ASK_USER>
<FINAL>...</FINAL>

Use only the public action contracts and candidate IDs supplied below.
\end{Verbatim}

Under native function calling, the equivalent wrapper uses exactly one supplied
tool call per turn and a dedicated route-submission action. A malformed block or
tool call consumes a turn; only a validator-accepted final route ends successfully.

\paragraph{Hermes.}
Hermes receives the same candidate boundary but a routing-specific coverage
contract: decompose the request into atomic needs, find the smallest supported
database set, and cite grounded evidence for every selected database.  The reference prompt gives a 30-turn budget and recommends the following schedule: turns 1--18 search/inspect, turns 19--23 finalize
coverage and topology, turn 24 should submit, and turns 25--30 are reserved for
validator-guided correction.  A valid submission terminates early.

\begin{table}[h]
\centering
\footnotesize
\caption{Hermes routing tools.}
\label{tab:app_hermes_tools}
\begin{tabular}{lp{0.61\linewidth}}
\toprule
Tool & Role \\
\midrule
\texttt{route\_search} & Search database, table, column, capability, and entity evidence. \\
\texttt{load\_database\_skill} & Load the public routing card for one opaque candidate. \\
\texttt{inspect\_database} & Inspect grounded schema/capability details. \\
\texttt{inspect\_relationships} & Inspect cross-source relations without treating compatibility as dependency. \\
\texttt{submit\_route} & Submit selected databases, one or more evidence items per database, grounded tables/columns/capabilities, database-local requests, and topology. \\
\bottomrule
\end{tabular}
\end{table}

\subsection{Planner and Interaction Prompts}
\label{app:planner_interaction_prompts}

The legacy LLM planner uses Chinese control instructions around English requests; its faithful English translation is reproduced below. The corrected interaction policy and semantic-slot judge are documented separately in Appendix~\ref{app:interactive_corrected}. Legacy interaction prompts are not treated as the implementation of the corrected run.

\paragraph{LLM Planner (faithful translation).}

\begin{Verbatim}[fontsize=\scriptsize,breaklines=true,breakanywhere=true]
SYSTEM: You are a professional database-query planning expert, skilled at
analyzing execution order and dependencies in multi-database queries. Output
JSON only.

TASK: Analyze the multi-database query below and determine database execution
order and dependencies.
USER QUERY: <query>
DATABASES TO ACCESS: <selected database list>

REQUIREMENTS:
1. Analyze dependencies among databases.
2. Determine the correct execution order (a topological ordering).
3. Return JSON:
{"execution_order": ["db1", "db2"],
 "dependencies": [{"from": "db1", "to": "db2",
   "type": "strong dependency", "reason": "db2 needs db1's result"}],
 "reason": "overall planning rationale"}
If no clear dependency exists, any order is allowed and dependencies is empty.
\end{Verbatim}

The call retries at most three times.  A parse failure falls back to the input
database order and scores as an unsuccessful planner call; any omitted selected
database is appended and any non-selected database is filtered.

\paragraph{Ask Once (corrected formal policy).}
The formal handler reads the canonical \texttt{ambiguity\_contract}. An outcome-blind decision precedes branch release; exact-match or semantic-slot alignment gates the natural-language answer, followed by actual rerouting and updated execution requests. The decision is shared across the two latent branch episodes. Appendix~\ref{app:interactive_corrected} defines scoring and information boundaries; Appendix~\ref{app:source_recheck} identifies the inspected source.

\subsection{Downstream SQL Generation}
\label{app:sql_prompts}

\begin{table}[h]
\centering
\footnotesize
\caption{Downstream prompt families and execution policy.}
\label{tab:app_sql_prompt_families}
\begin{tabular}{p{0.20\linewidth}p{0.44\linewidth}p{0.26\linewidth}}
\toprule
Corpus/agent & Prompt content & Output and stopping rule \\
\midrule
Spider Text-to-SQL & SQLite DDL and comments followed by the database-local request and ``Generate the SQL after thinking step by step:'' & Read-only SQL parsed after a seeded \texttt{SELECT}; fail closed on invalid output. \\
BIRD Text-to-SQL & SQLite schema, external-knowledge block when available, database-local request, and the same generation suffix & Temperature 0; read-only SQL; execution/equivalence scorer. \\
LiveSQLBench Text-to-SQL & PostgreSQL schema, column meanings, external knowledge, and task & Exactly one read-only query in a \texttt{postgresql} code block. \\
Dataset-specific adapter & Frozen DAIL-SQL/ScaleSQL-style prompt and stage configuration appropriate to the upstream corpus & Same source-unit execution boundary and denominator as generic Text-to-SQL. \\
LiveSQLBench agent & PostgreSQL expert instruction with schema/search/query/test/submit tools & At most 30 steps, test before one final submission, and no write operation. \\
\bottomrule
\end{tabular}
\end{table}

Source execution is configured to use read-only controls.  A source-level parser,
generation, timeout, or execution failure yields an incorrect source unit and
remains in the task denominator; it is not replaced with gold SQL.  Sequential
tasks substitute only validated upstream outputs into declared bindings.

\section{Extended Results and Failure Analysis}
\label{app:extended_results}

The complete machine-readable routing result contains 42 model--router rows,
and the homogeneous configuration table contains 288 model--configuration rows.
These are repeated evaluations of the same task set, not independent experimental
samples. The main paper reports the full-815 matrices, with a separate non-interactive sensitivity analysis.

\subsection{Routing Quality and Failure Direction}

Under the evaluated method-specific schedules, Hermes has the highest recorded
RA-Set and RA-Graph row for each backend in the full 815-record matrix. This is
a descriptive property of those evaluated conditions, not a compute- or evidence-matched
algorithmic comparison. Directional errors and the full rows are preserved below
for auditability.

This decomposition also separates retriever-limited and agent-limited failures.
Candidate Top-50 already lacks at least one required database for 13.50\% of
formal tasks.  When a required source is available but still absent from the
submitted route, the failure lies in evidence acquisition, source retention, or
route commitment rather than in the catalog substrate alone.

\subsection{Database-Level Query Representation}

Privileged source questions achieve the highest observed source-execution scores, but they are not a clean intent-preservation intervention: selected frozen cards show that evaluator-side source questions can contain constraints not explicit in the visible request. The main paper therefore interprets the gap as a mixture of decomposition loss and specification completeness. The full-815 fixed LLM/Never Ask diagnostic is shown in Figure~\ref{fig:querymode}; its matched-effect tables appear in Appendix~\ref{app:numeric}.

\subsection{Sequential Planning}
\label{app:sequential_planning}

\begin{figure}[h]
\centering
\includegraphics[width=\linewidth]{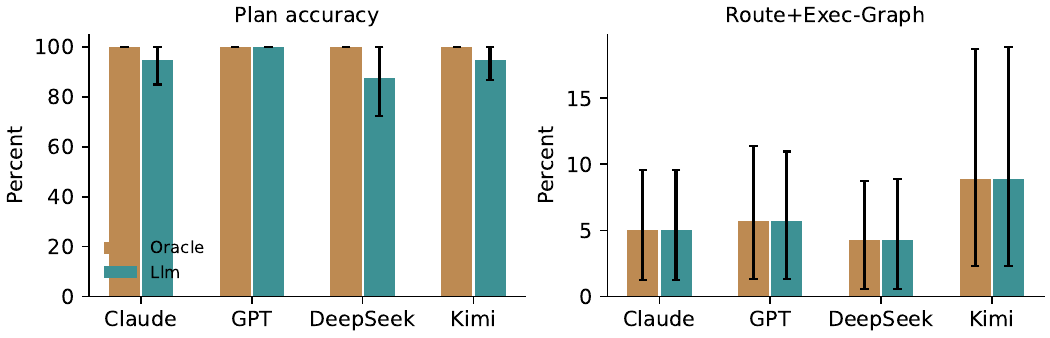}
\caption{Oracle versus LLM order on correctly routed Sequential tasks. Near-ceiling plan accuracy coexists with a low route+execution conjunction score; each backend has only 19--20 distinct tasks in this conditional analysis.}
\label{fig:planner}
\end{figure}

When routing is already correct, the LLM planner approaches the Oracle planner
on Sequential tasks while the route+execution conjunction remains much lower. The conditional
subsets contain only 19--20 distinct tasks per backend. LLM plan accuracy is
87.5\%--100\%, while conditional Route+Exec-Graph is 4.31\%--8.92\% and is unchanged
by Oracle order in these evaluations. The absence of a conjunction-score gain from this order
intervention should not be generalized to long-horizon planning or more complex
dependency graphs.

\subsection{Interactive Episodes and Branch-Weighted Aggregation}
\label{app:interactive_corrected}
Each of the 105 original Interactive tasks is evaluated on two latent-intent episodes with weight 0.5 per branch. Ordinary tasks retain unit weight. Each configuration therefore has 920 episode-level records (710 ordinary plus 210 branch episodes), collapsed to 815 base-task rows. The 288-cell publication table therefore contains 234,720 task rows derived from 264,960 physical records. The original task identities and non-interactive results are retained.

\paragraph{Information boundary and response release.}
An outcome-blind agent fixes whether to ask and the clarification question before either latent answer is exposed. The same decision is reused across both episodes. A question must identify the canonical user-facing semantic slot. Exact normalized matches are accepted; otherwise the formal handler uses a semantic alignment judge. Content-free placeholders, incompatible slots, and explicit alternatives that exclude the canonical values fail closed. An open question need not enumerate hidden branch labels. Only a valid canonical natural-language answer is released; targets, source-local gold, and SQL remain evaluator-private.

\paragraph{Routing and execution.}
After response release, the same routing policy processes the clarified request. Its revised source set and local requests feed downstream execution. Global fused execution receives the visible request plus clarification, without routing-only instructions. Router local uses the revised local requests. Privileged source mode separately uses the evaluator's source-local question. Never Ask receives no latent answer: its unclarified prediction is evaluated against both possible singleton targets under the same prior. The runtime scores the branch's target and gold even if the response was not released; it does not silently revert to the union of alternatives.

\paragraph{Structural gate and statistical unit.}
Interactive RA-Graph requires exact branch-source selection, normalized topology \texttt{interactive}, and an empty dependency-edge set. A singleton route normalized as \texttt{single\_target} does not pass this label gate. The score therefore tests both source recovery and identification of the interaction regime. For each branch, all conjunctions are computed before averaging: $\sum_b w_b g_bt_bh_b$, not the product of separately averaged gates. CEA uses the sum of branch-level conjunction numerators divided by the corresponding summed routing gate. Each source-count denominator also follows the selected branch; every full cell has 2,293 weighted required contributions, rather than the historical union denominator of 2,398.

The main matched comparison fixes LLM planning and non-privileged Fused/Router local inputs, averaging 12 cells per policy. We resample the 105 base tasks 5,000 times, retaining both branches and every matched cell together. The uniform two-branch prior is an evaluation choice, not an estimated distribution of real user intent. These intervals quantify task-sampling uncertainty conditional on fixed experiment outputs and do not capture provider/seed, simulator-judge, or all shared-database variation.

\input{appendix/interactive_corrected}

On this slice, Never Ask TEA is 6.94\%--9.33\%, while RA-Graph is only 0--0.16\% and Route+Exec-Graph is zero. Its low conjunction is therefore partly a behavior-label gate, not evidence that every generated SQL query is incorrect. Ask Once reaches 35.91\%--38.02\% Route+Exec-Graph. The source-set gains of 39.84--43.17 percentage points are accompanied by nonzero repair and occasional harm (Appendix~\ref{app:numeric}). These results establish improvement under this canonical single-clarification protocol; they do not validate arbitrary multi-turn dialogue, human preference satisfaction, or final-answer synthesis.

\subsection{Non-Interactive Sensitivity Analysis and Transition Rates}
\label{app:noninteractive_sensitivity}

To assess the contribution of the Interactive regime, we separately aggregate the 710 Single-target, Parallel, Sequential, and Exploratory tasks. This post-hoc sensitivity slice was not pre-registered. It makes zero new model calls and changes no ordinary task, route, SQL verdict, or comparator. All 815 base tasks remain in the corrected headline scores. The factorial maximum below is specifically the Never Ask non-privileged control; it is not the larger Ask Once maximum.

\begin{table}[h]
\centering
\small
\caption{Full-815 headline scores versus the 710-task non-interactive sensitivity slice. The latter is a post-hoc analysis, not a pre-registered benchmark split. Maximum rows are descriptive and are not used as the main factorial summary.}
\label{tab:noninteractive_sensitivity}
\begin{tabular}{lrr}
\toprule
Quantity & Main all-815 & Sensitivity 710 \\
\midrule
Highest routing RA-Set & 60.55\% & 66.20\% \\
Highest routing RA-Graph & 34.66\% & 39.72\% \\
Highest Never Ask non-privileged Route+Exec & 11.53\% & 13.24\% \\
\bottomrule
\end{tabular}
\end{table}

The nested score ordering alone is not evidence of a new phenomenon, because
RA-Graph contains RA-Set correctness and Route+Exec-Graph contains RA-Graph.
A more informative diagnostic is the conditional transition rate on one fixed
non-oracle reference slice (Hermes + corpus-specific adapter + LLM planner +
Never Ask + Router local):

\begin{table}[h]
\centering
\small
\caption{Conditional transitions on one fixed non-oracle 710-task reference slice. These rates quantify failure mass between nested gates on the same evaluated tasks; they do not make the reference router compute matched to other policies.}
\label{tab:noninteractive_transitions}
\begin{tabular}{lrrr}
\toprule
Backend & RA-Set & $P(\mathrm{RA\mbox{-}Graph}\mid\mathrm{RA\mbox{-}Set})$ & $P(\mathrm{Route{+}Exec}\mid\mathrm{RA\mbox{-}Graph})$ \\
\midrule
Claude Opus 5 & 66.34\% & 59.87\% & 33.33\% \\
GPT-5.6 Sol & 65.92\% & 57.48\% & 32.34\% \\
DeepSeek-V4-Flash & 56.62\% & 58.46\% & 37.02\% \\
Kimi K3 & 61.13\% & 58.29\% & 33.60\% \\
\bottomrule
\end{tabular}
\end{table}

This table supports the narrower diagnostic claim that, even after conditioning
on the prior gate, a substantial fraction of evaluated tasks fail the next measured
gate. It does not isolate behavior-label, topology, and dependency-edge errors
within RA-Graph; the available aggregate results do not support that decomposition.

\subsection{Descriptive Best-of-Grid Results}
\label{app:best72}
The appendix reports the per-model Best-of-72 rows and
Top-4 configuration summaries for completeness. They are evaluation-set-selected
descriptive maxima, include privileged local-question conditions, and receive no
selection-corrected uncertainty interval. For that reason they are not used as
main-text method rankings or headline capability estimates in the revised reporting.

\subsection{Stage-Wise Failure Taxonomy}
\label{app:failure_taxonomy}

\paragraph{Retrieval-substrate miss.}
At least one required database is absent from the exposed candidate/retrieved
set.  This is a property of AllTargetCoverage@K, not agent route accuracy.

\paragraph{Agent under-selection.}
A required database is available under the observable evidence boundary but is
omitted from the submitted route.

\paragraph{Agent over-selection.}
The route contains unnecessary databases.  A system can preserve high recall
while still failing exact-set correctness.

\paragraph{Route-graph error.}
The exact database set is correct but topology or dependency structure is
incorrect.

\paragraph{Sequential binding/order error.}
The correct sources are selected but a verified output-to-input binding or
required execution order is not preserved.

\paragraph{Interactive branch error.}
The system fails to resolve a genuine ambiguity to the correct executable
branch.

\paragraph{Intent-decomposition error.}
The correct database is selected, but its database-local request does not
preserve that source's role in the global task.

\paragraph{Source-execution error.}
The route is structurally correct but one or more required source-level
executions are incorrect.

\paragraph{Interaction without belief revision.}
The system asks for clarification but does not update the effective source
hypothesis after receiving the answer.

\section{Task Examples and Validity Cases}
\label{app:examples}

The following frozen cards serve two purposes: the Single-target and Sequential
cards make the primary routing mechanics concrete, while the Parallel,
Exploratory, and Interactive cards expose additional structural regimes and
published validity boundaries. They are therefore diagnostic examples, not a claim that
every selected wording is a quality exemplar. We use
$|\mathcal{C}_{\mathrm{task}}|$ for the size of the complete task-level
candidate database catalog. This is distinct from a RouteEnv retrieved
Top-$K$ pool: the former defines the databases available in the benchmark
instance, whereas the latter is an agent observation produced by retrieval.
For exposition only, the cards below reveal evaluator-side physical database
names and structural gold annotations. The evaluated router receives the
frozen user request, the candidate environment, public RouteEnv observations,
and opaque \texttt{db\_*} identifiers; it is not shown the gold source names,
source-local decomposition, dependency bindings, or branch labels reported
below.

\begin{center}
\scriptsize
\begin{tabular}{lllrrp{4.2cm}}
\toprule
Behavior & Corpus & Diff. & $|\mathcal{C}_{\mathrm{task}}|$ & $|G|$ & Structural feature \\
\midrule
Single-target & LiveSQLBench & Med. & 11 & 1 & Single-source discovery control \\
Parallel & BIRD & Med. & 16 & 2 & Independent multi-source evidence \\
Sequential & Spider & Med. & 22 & 2 & Upstream-output-to-downstream-input binding \\
Exploratory & BIRD & Hard & 80 & 6 & Capability discovery over a corpus-scale catalog \\
Interactive & Spider & Med. & 97 & 2 & Clarification-conditioned source routing \\
\bottomrule
\end{tabular}
\end{center}

\subsection{Single-target: Source Discovery without Cross-source Structure}
\label{app:example_single}

\noindent\textbf{Example setting.} LiveSQLBench; $|\mathcal{C}_{\mathrm{task}}|=11$, $|G|=1$.

\noindent\textbf{Frozen user request.}
\begin{quote}\small
How much do we typically recover through each return method? Give me the
average amounts rounded to pennies.
\end{quote}

\noindent\textbf{Gold source-local intent.}
The unique required source is \nolinkurl{reverse_logistics_large}.  Its local
query groups returns by return channel and computes the average recovery value,
rounded to two decimal places.

\noindent\textbf{Workflow structure.}
There is no dependency edge and no clarification branch.  This task therefore
acts as a control that isolates whether the router can discover the correct
source from distractors before any multi-source planning is required.

\subsection{Parallel Example: Two Independent Sources under a Shared Metric}
\label{app:example_parallel}

\noindent\textbf{Example setting.} BIRD; $|\mathcal{C}_{\mathrm{task}}|=16$, $|G|=2$.

\noindent\textbf{Frozen user request.}
\begin{quote}\small
Prepare one comparative report with independent sections. Address each request
exactly as stated: (1) Who among the players in season 2000-2001 has committed
the highest rule violations or penalty minutes? (2) Between 2003 to 2005, what
are the given names of the players with the most number of games played whose
Penalty minutes is between 200 to 250? Keep every result tied to its original
metric, filter, condition, time frame, grouping key, requested field, and unit.
\end{quote}

\noindent\textbf{Gold source set and local intents.}
The required sources are \nolinkurl{ice_hockey_draft}, which identifies the
2000--2001 player with the highest penalty-minute measure, and
\nolinkurl{hockey}, which returns the requested player names for 2003--2005
under the 200--250 penalty-minute constraint.

\noindent\textbf{Workflow structure.}
The two source units are independent: neither consumes an output produced by
the other. The construction annotation associates both results with a
penalty-minute theme. Route+Exec-Graph verifies the two required source
executions rather than independently scoring the prose-level comparative
synthesis. This card therefore isolates \emph{multi-source completeness} from
sequential dependency reasoning: a system must recover both sources even
though either source can be executed first.

\subsection{Sequential: A Hidden Cross-database Binding}
\label{app:example_sequential}

\noindent\textbf{Example setting.} Spider; $|\mathcal{C}_{\mathrm{task}}|=22$, $|G|=2$.

\noindent\textbf{Frozen user request.}
\begin{quote}\small
Find the types of vocals used in the song with the same title as a film that is
rented at a fee of 0.99 and has less than 3 in inventory.
\end{quote}

\noindent\textbf{Gold source-local decomposition.}
The first source, \nolinkurl{sakila_1}, identifies the qualifying film and emits
its \texttt{title}.  The second source, \nolinkurl{music_2}, retrieves vocal
types for a song whose title equals that upstream result.

\noindent\textbf{Dependency and binding.}
The frozen order is \nolinkurl{sakila_1} $\rightarrow$
\nolinkurl{music_2}, with the first source's \texttt{title} binding the
second-source placeholder \texttt{:film.title}.  The evaluator-side gold
execution resolves this value to \texttt{Badlands}; critically, that literal is
\emph{gold-only} and is not present in the frozen user request or router-visible
input.  Without the upstream output, the downstream request is not fully
instantiable.  This is the core distinction between DSABench sequential tasks
and merely concatenating two independent Text-to-SQL questions.

\subsection{Exploratory Validity Case: Capability Discovery over the Full BIRD Catalog}
\label{app:example_exploratory}

\noindent\textbf{Example setting.} BIRD; $|\mathcal{C}_{\mathrm{task}}|=80$, $|G|=6$.

\noindent\textbf{Frozen user request.}
\begin{quote}\small
Which registered data sources are needed to evaluate titles of works filtered
by domain-specific criteria? Use the discovered evidence to relate recipe
titles filtered by nutritional criteria and film titles rated for general
audiences; movie list titles for user 32172230 and movie titles before 2000;
and the titles of the papers published in the journal ``Concepts in Magnetic
Resonance Part A'' in 2008 and movie titles by Jack Kinney. Preserve every
metric, filter, time frame, and unit in the synthesis.
\end{quote}

\noindent\textbf{Gold capability-to-source mapping.}
The six required sources are \nolinkurl{cookbook} (recipe titles under the
specified nutritional criterion), \nolinkurl{movie_3} (general-audience film
titles), \nolinkurl{movie_platform} (trial-eligible movie-list titles for user
32172230), \nolinkurl{movies_4} (five movie titles released before 2000),
\nolinkurl{authors} (2008 paper titles in the named journal), and
\nolinkurl{disney} (movie titles directed by Jack Kinney).

\noindent\textbf{Workflow structure.}
The task first requires capability discovery across all 80 BIRD databases and
then resolves to six independent source units followed by a multi-aspect
summary. No database name is supplied in the user request. This card also
serves as a validity example: the visible request does not state every
source-local constraint present in executable gold. The exact mismatches are
listed in Section~\ref{app:example_gold}; the record is preserved because it is
part of the evaluated frozen snapshot rather than presented as a fully
specified language exemplar.

\subsection{Interactive Validity Case: Clarification Changes the Required Source}
\label{app:example_interactive}

\noindent\textbf{Example setting.} Spider; $|\mathcal{C}_{\mathrm{task}}|=97$, two possible gold branches.

\noindent\textbf{Frozen underspecified request.}
\begin{quote}\small
How many courses are offered by the department?
\end{quote}

\noindent\textbf{Released clarification.}
\begin{quote}\small
Which department do you mean: Physics or Computer Info. Systems?
\end{quote}

\noindent\textbf{Gold branches.}
Physics resolves to \nolinkurl{college_2} and its distinct-course query; Computer Info. Systems resolves to \nolinkurl{college_1} and its course-count query. The corrected evaluator scores each target and local gold in a separate half-weight episode. The router does not see either private target mapping before response release. The canonical question shown above is evaluator reference text, not necessarily the question generated by the agent.

\noindent\textbf{Why interaction is necessary.}
The initial request does not identify the department, and the two interpretations require different sources. A released answer therefore changes the applicable target and local query. The controlled protocol tests whether the agent asks about the department slot and uses the reply to recover the corresponding source and execution.

\subsection{Executable Gold and a Concrete Binding}
\label{app:example_gold}
These snippets are evaluator-side examples copied from the frozen task set, not inputs supplied to the evaluated router. The sequential downstream gold SQL is:
\begin{Verbatim}[fontsize=\scriptsize,breaklines=true]
SELECT TYPE FROM vocals AS T1 JOIN songs AS T2
ON T1.songid = T2.songid WHERE title = "Badlands"
\end{Verbatim}
The corresponding contract binds upstream field \texttt{title} to parameter \texttt{:film.title} and replaces the downstream gold literal with the validated upstream result. The literal \texttt{Badlands} belongs to executable gold; it is not present in the visible sequential request. Correct source selection alone does not supply that value.

The single-target PostgreSQL example uses:
\begin{Verbatim}[fontsize=\scriptsize,breaklines=true]
SELECT r.return_channel,
       ROUND(AVG((fm.cost_breakdown->'valuation'->>
                  'recovery_value')::numeric),2) AS avg_recovery_value
FROM returns r
JOIN financial_management fm ON fm.casetag = r.casenum
GROUP BY r.return_channel
\end{Verbatim}
For the exploratory card, two gold constraints absent from the visible request are \texttt{user\_eligible\_for\_trial = 1} and \texttt{LIMIT 5}; the recipe query additionally uses \texttt{iron > 20}, which is explicitly grounded in the BIRD source unit's external evidence. The latter is an evidence-visibility question, not an unsupported gold predicate. The visible fused request still omits the trial condition and five-title cardinality that appear in the source questions. This concrete contrast explains why the oracle-question gap can combine decomposition difficulty with underspecification, rather than identifying a purely model-side failure.

\section{Reproducibility Resources and Release Scope}
\label{app:reproducibility}

The appendix summarizes the finalized benchmark and recorded experimental outputs; preparing these summaries did not involve rerunning models. Reproducing the experiments themselves additionally requires the original databases, model endpoints, prompts, and generated SQL. The release provides:
\begin{itemize}
  \item the finalized 815-task benchmark definition;
  \item task profiles, target/candidate distributions, and registry summaries;
  \item source-corpus, behavior, and difficulty summaries;
  \item adjudication decisions and question-text repair records;
  \item the seven-model routing evaluation matrix;
  \item renderer examples, prompt templates, public tool schemas, validators, retry limits, and stopping rules;
  \item 42 routing summaries with dataset and behavior breakdowns;
  \item 288 homogeneous configuration summaries and task-level outputs;
  \item matched request-mode, planner, clarification, and factorial analyses;
  \item Best-of-72 and Top-4 configuration summaries;
  \item RouteEnv source-unit and formal-task retrieval metrics;
  \item plotting code and environment specifications; and
  \item scripts for checking the aggregate statistics.
\end{itemize}

The distribution distinguishes the full 303-name/304-instance registry from the 263-name/264-instance subset appearing in finalized tasks. Private identity mappings, API keys, provider credentials, and other evaluation secrets are not part of the public release.

\subsection{Modular Evaluation Components}

Benchmark construction and validation, human review, RouteDB/RouteEnv preparation, routing evaluation, combined routing-and-execution evaluation, and metric aggregation are separate components. This separation lets new routers and executors be tested on the same task definitions without changing the gold semantics.

\subsection{Technical Failures}

Task counts are determined before scoring. Provider errors, parser failures, tool errors, and missing execution outputs remain in the evaluation denominator under the stated rules. Excluding such cases after the fact would make systems with higher failure rates appear artificially strong.

\section{Benchmark Datasheet, Responsible Release, and Validity}
\label{app:datasheet}

\subsection{Benchmark Datasheet}

\paragraph{Motivation.}
DSABench studies the source-selection and multi-database planning decisions that standard database-given Text-to-SQL tasks make in advance.

\paragraph{Composition.}
The benchmark has 815 reviewed tasks built from executable units in Spider, BIRD, and LiveSQLBench. All tasks contribute to the primary evaluation; the 710 non-interactive tasks are also analyzed separately. Each task specifies a candidate environment and evaluator-private source targets, local queries, behavior annotations, and executable structure.

\paragraph{Collection and creation.}
DSABench does not collect private user data. We turn existing research datasets into source units, compose them using fixed executable contracts, generate user-facing requests, check them automatically, and finalize them through two-pass human review and independent adjudication.

\paragraph{Preprocessing and annotations.}
Database identities are scoped to their source dataset, and source SQL is read-only. Routers see public catalog metadata rather than private database mappings. Gold source sets, route structures, SQL, behavior labels, and review decisions are benchmark annotations, not predictions produced by the evaluated agent.

\paragraph{Recommended uses.}
Database routing, tool-use policies, exact multi-source discovery, joint
route-structure evaluation, database-local request decomposition, and staged
required-source execution diagnostics. Interactive records support controlled two-branch source-and-execution evaluation under the explicit corrected protocol; unrestricted dialogue quality remains out of scope.

\paragraph{Out-of-scope uses.}
A benchmark score should not be treated as evidence that a system meets production authorization, security, privacy, fairness, or safety requirements, particularly in high-stakes settings.

\paragraph{Distribution.}
Licensing is component-wise. Upstream database contents remain governed by
their original terms and are not relicensed by DSABench; users must obtain
upstream assets separately where redistribution is not permitted. A public
reuse release must pair the DSABench code/annotation terms with an explicit
upstream attribution and redistribution documentation. The DSABench release does not grant rights to redistribute upstream databases. Evaluator secrets and provider credentials are excluded.

\subsection{Maintenance and Versioning}
\label{app:maintenance}

The reported results use a fixed set of 815 tasks, database identities, evaluator rules, and metric definitions. Future corrections should be published as explicit benchmark revisions with a record of changes, not made silently. Wording-only repairs should be distinguishable from changes that affect a task's executable meaning.

\subsection{Threats to Validity}
\label{app:validity}

\paragraph{Corpus representativeness.}
The catalog is larger than the setting of typical database-given Text-to-SQL tasks but remains smaller and more orderly than many enterprise data estates.

\paragraph{Underlying-corpus exposure.}
Because the upstream datasets are public, their schemas may have appeared during model pretraining. Our results concern multi-source discovery and workflow construction over these datasets, not generalization to entirely unseen database contents.

\paragraph{Single-call retrieval coverage.}
RouteEnv is deliberately non-oracle.  Candidate AllTargetCoverage@50 is 86.50\%
on formal tasks; this limits a policy that commits only from that one retrieved set. Adaptive agents can issue different searches or inspect additional allowed candidates, so it is not an upper bound on their possible accuracy.

\paragraph{Strict exactness.}
RA-Set and RA-Graph require exact matches and therefore do not credit routes that are only partly correct. We also report precision, recall, and under- and over-selection errors to make that partial progress visible.

\paragraph{Behavior coverage.}
Only 22 tasks test explicit Sequential value-flow dependencies. Interactive results cover 105 tasks but rely on a two-branch simulator and an alignment judge, limiting what they say about open-ended conversation.

\paragraph{Surface-form cues.}
As quantified in Section~\ref{app:surface_audit}, standardized source-discovery
wording appears in much of the frozen Exploratory split.  Systems can therefore
partly infer the behavior family from wording.  The reported metrics remain well-defined for the
published snapshot but should not be interpreted as evidence of robustness to
paraphrase or naturally occurring exploratory requests.

\paragraph{Control-prompt language.}
The legacy planner uses Chinese instructions around English requests; the corrected semantic-slot alignment judge uses English instructions. These implementation choices can affect absolute performance. Matched configurations preserve their corresponding control prompts; language robustness is not separately measured.

\paragraph{Human validation.}
Human review and adjudication are safeguards, not guarantees that every request has a unique interpretation. We know that 612 candidates received one review and 744 received two, but the available records do not support a reliable agreement estimate or a reconstruction of review coverage for the final 815. Some retained examples also reveal differences between user wording and source-local requirements; we do not extrapolate an all-task mismatch rate from them.

\paragraph{Grid-selection interpretation.}
Evaluation-set-selected maxima are reported in the appendix for completeness,
but the main paper reports the non-oracle score distribution rather than using
Best-of-72 as a method ranking.

\subsection{Evaluation Scope}
The main results use all 815 tasks and the corrected branch-conditioned Interactive protocol. We also report a post hoc 710-task analysis without Interactive tasks. Route+Exec-Graph combines exact source selection, route structure, required-source execution, and sequential planning where applicable; it does not independently grade the final natural-language answer. Earlier union-list interaction scores are not equivalent to scores under the corrected protocol.

\input{appendix/numeric_tables}

\FloatBarrier
\section{Validation of the Interactive Evaluation Protocol}
\label{app:source_recheck}
We checked that the Interactive evaluator aligns clarification questions to predefined semantic slots, makes the ask decision before revealing the hidden interpretation, returns only the permitted branch answer, and distinguishes routing context from the clarified request used for execution. Each configuration must contain 710 ordinary tasks and two episodes for each of 105 Interactive task identities. Validation also checks branch consistency and that both episodes share the same Ask Once decision.

The seven-backend routing study is rescored separately: its original route outputs are judged against the two single-branch targets, with the required \texttt{interactive} label and empty dependency edges. This procedure does not use the newer Ask Once outputs or perform clarification. Routing-only and factorial values therefore need not match.

Independent checks cover the 288 factorial configurations, verify the 815 base-task identities in each, and confirm the branch-weighted aggregates. Reproducing model inference also requires the original prompts, responses, SQL, databases, and compatible endpoints.

\paragraph{Evaluation-cost scope.}
The corrected Interactive experiments include clarification alignment, rerouting, and downstream generation. We do not report cross-provider efficiency rankings because endpoint accounting conventions are not directly comparable.

%% file: appendix/interactive_corrected.tex
\begin{table}[!htbp]
\centering\scriptsize
\caption{Matched Interactive results on 105 base tasks. LLM planning and non-privileged inputs are fixed; 12 cells are averaged per policy. Branches receive equal weight; 95\% intervals use 5,000 base-task cluster resamples. Never Ask has zero conjunction score, so Ask conjunction also equals its gain.}
\label{tab:interactive_corrected}
\begin{tabular}{lrrrrr}\toprule
Backend & Set Never & Set Ask & $\Delta$ Set [CI] & R+E Ask & R+E CI \\
\midrule
Claude Opus 5 & 18.25 & 61.43 & 43.17 [36.51, 49.52] & 38.02 & [31.94, 44.05] \\
GPT-5.6 Sol & 19.84 & 60.32 & 40.48 [34.92, 46.19] & 35.91 & [30.48, 41.35] \\
DeepSeek-V4-Flash & 17.78 & 58.25 & 40.48 [35.08, 45.71] & 36.79 & [30.95, 42.54] \\
Kimi K3 & 18.41 & 58.25 & 39.84 [34.13, 45.40] & 36.23 & [30.79, 41.91] \\
\bottomrule\end{tabular}
\end{table}

\paragraph{Fixed-control repair accounting.}
Using the same LLM-planner, non-privileged 12-cell control, the ask rates are 100.00\%, 99.37\%, 97.14\%, and 94.60\% for Claude, GPT, DeepSeek, and Kimi. Among asked observations, source-set repair rates are 45.71\%, 43.13\%, 43.30\%, and 44.63\%; harm rates are 2.54\%, 2.40\%, 1.63\%, and 2.52\%. Repair changes an incorrect exact set into a correct one; harm reverses that transition. These conditional rates divide pooled half-weight event counts by pooled half-weight ask counts across the fixed cells. They are not task-macro conditional means or the all-grid marginal rates. The denominator and control match the main-text mechanism interpretation; the accompanying derived table preserves unrounded values.

%% file: appendix/numeric_tables.tex
\section{Complete Numeric Supplements}
\label{app:numeric}
The tables summarize the corrected evaluation outputs. Routing-only and factorial rows come from distinct evaluations and should not be interpreted as paired observations unless explicitly stated.
\begin{table}[!htbp]
\centering\scriptsize
\caption{Routing-only: Claude Opus 5. 815 base tasks; percent.}
\label{tab:routing_updated_0}
\begin{tabular}{lrrrrrr}\toprule
Policy & RA-Set & RA-Graph & F1 & OR & UR & Valid \\
\midrule
Random & 1.47 & 0.86 & 3.88 & 93.99 & 96.85 & 100.00 \\
Rerank & 24.36 & 17.55 & 51.14 & 18.43 & 53.91 & 81.10 \\
Direct & 24.11 & 17.18 & 53.84 & 19.66 & 51.89 & 87.36 \\
Constrained & 22.88 & 16.93 & 49.75 & 19.48 & 55.60 & 81.35 \\
ReAct & 39.45 & 25.03 & 51.19 & 12.00 & 50.81 & 67.48 \\
Hermes & 60.06 & 34.66 & 81.29 & 14.04 & 19.97 & 99.39 \\
\bottomrule\end{tabular}
\end{table}
\begin{table}[!htbp]
\centering\scriptsize
\caption{Routing-only: GPT-5.6 Sol. 815 base tasks; percent.}
\label{tab:routing_updated_1}
\begin{tabular}{lrrrrrr}\toprule
Policy & RA-Set & RA-Graph & F1 & OR & UR & Valid \\
\midrule
Random & 1.47 & 0.86 & 3.88 & 93.99 & 96.85 & 100.00 \\
Rerank & 23.56 & 16.56 & 50.82 & 19.62 & 55.27 & 84.66 \\
Direct & 23.74 & 16.32 & 53.61 & 19.44 & 53.21 & 88.83 \\
Constrained & 24.17 & 16.93 & 52.63 & 18.53 & 53.74 & 85.89 \\
ReAct & 49.82 & 30.67 & 74.09 & 16.07 & 29.30 & 96.32 \\
Hermes & 60.55 & 33.01 & 81.50 & 14.36 & 20.64 & 100.00 \\
\bottomrule\end{tabular}
\end{table}
\begin{table}[!htbp]
\centering\scriptsize
\caption{Routing-only: Kimi K3. 815 base tasks; percent.}
\label{tab:routing_updated_2}
\begin{tabular}{lrrrrrr}\toprule
Policy & RA-Set & RA-Graph & F1 & OR & UR & Valid \\
\midrule
Random & 1.47 & 0.86 & 3.88 & 93.99 & 96.85 & 100.00 \\
Rerank & 21.23 & 15.34 & 35.65 & 11.51 & 67.31 & 53.87 \\
Direct & 21.78 & 15.83 & 46.74 & 17.47 & 58.40 & 76.32 \\
Constrained & 20.25 & 14.85 & 38.28 & 12.84 & 65.30 & 59.39 \\
ReAct & 52.15 & 30.31 & 70.32 & 14.64 & 32.05 & 89.45 \\
Hermes & 55.83 & 30.92 & 79.03 & 15.35 & 22.94 & 99.26 \\
\bottomrule\end{tabular}
\end{table}
\begin{table}[!htbp]
\centering\scriptsize
\caption{Routing-only: DeepSeek-V4-Pro. 815 base tasks; percent.}
\label{tab:routing_updated_3}
\begin{tabular}{lrrrrrr}\toprule
Policy & RA-Set & RA-Graph & F1 & OR & UR & Valid \\
\midrule
Random & 1.47 & 0.86 & 3.88 & 93.99 & 96.85 & 100.00 \\
Rerank & 18.96 & 13.25 & 34.84 & 11.34 & 68.71 & 54.48 \\
Direct & 18.40 & 12.88 & 33.32 & 11.85 & 69.93 & 52.76 \\
Constrained & 19.08 & 13.62 & 35.52 & 11.23 & 68.10 & 54.97 \\
ReAct & 42.02 & 26.38 & 61.29 & 15.11 & 42.07 & 83.07 \\
Hermes & 50.06 & 27.48 & 76.96 & 15.71 & 26.39 & 99.88 \\
\bottomrule\end{tabular}
\end{table}
\begin{table}[!htbp]
\centering\scriptsize
\caption{Routing-only: DeepSeek-V4-Flash. 815 base tasks; percent.}
\label{tab:routing_updated_4}
\begin{tabular}{lrrrrrr}\toprule
Policy & RA-Set & RA-Graph & F1 & OR & UR & Valid \\
\midrule
Random & 1.47 & 0.86 & 3.88 & 93.99 & 96.85 & 100.00 \\
Rerank & 22.52 & 16.20 & 51.11 & 16.86 & 55.14 & 82.45 \\
Direct & 23.25 & 16.44 & 51.26 & 16.50 & 55.02 & 82.45 \\
Constrained & 22.58 & 16.32 & 51.50 & 15.95 & 55.10 & 83.07 \\
ReAct & 29.75 & 18.65 & 45.96 & 12.31 & 57.64 & 66.50 \\
Hermes & 52.33 & 28.83 & 77.46 & 15.20 & 26.13 & 99.88 \\
\bottomrule\end{tabular}
\end{table}
\begin{table}[!htbp]
\centering\scriptsize
\caption{Routing-only: Qwen3.5-397B-A17B. 815 base tasks; percent.}
\label{tab:routing_updated_5}
\begin{tabular}{lrrrrrr}\toprule
Policy & RA-Set & RA-Graph & F1 & OR & UR & Valid \\
\midrule
Random & 1.47 & 0.86 & 3.88 & 93.99 & 96.85 & 100.00 \\
Rerank & 23.80 & 16.93 & 52.24 & 21.37 & 53.35 & 86.26 \\
Direct & 23.80 & 16.81 & 52.89 & 21.95 & 52.76 & 87.98 \\
Constrained & 24.23 & 16.93 & 52.70 & 20.43 & 52.88 & 86.13 \\
ReAct & 36.87 & 23.19 & 56.08 & 13.56 & 47.62 & 77.42 \\
Hermes & 48.34 & 27.12 & 72.70 & 15.03 & 30.56 & 94.11 \\
\bottomrule\end{tabular}
\end{table}
\begin{table}[!htbp]
\centering\scriptsize
\caption{Routing-only: Qwen3.6-35B-A3B. 815 base tasks; percent.}
\label{tab:routing_updated_6}
\begin{tabular}{lrrrrrr}\toprule
Policy & RA-Set & RA-Graph & F1 & OR & UR & Valid \\
\midrule
Random & 1.47 & 0.86 & 3.88 & 93.99 & 96.85 & 100.00 \\
Rerank & 17.55 & 12.76 & 33.76 & 11.03 & 69.53 & 52.64 \\
Direct & 18.04 & 13.25 & 33.63 & 11.95 & 69.57 & 53.25 \\
Constrained & 15.83 & 12.02 & 31.72 & 10.81 & 71.57 & 50.06 \\
ReAct & 37.18 & 23.44 & 62.80 & 16.67 & 41.76 & 89.20 \\
Hermes & 47.48 & 26.87 & 69.60 & 13.26 & 33.41 & 89.20 \\
\bottomrule\end{tabular}
\end{table}
\FloatBarrier
\begin{table}[!htbp]
\centering\scriptsize
\caption{Top four evaluated configurations: Claude Opus 5. Evaluation-set-selected descriptive maxima; privileged inputs are explicitly marked.}
\label{tab:top4_0}
\begin{tabular}{lrrrrrr}\toprule
Router & Executor & Planner & Policy & Input & EA & R+E \\
\midrule
Hermes & Adapter & Direct & Ask & Privileged & 63.69 & 24.79 \\
Hermes & Adapter & Oracle & Ask & Privileged & 63.78 & 24.79 \\
ReAct & Adapter & Direct & Ask & Privileged & 28.98 & 20.25 \\
ReAct & Adapter & Oracle & Ask & Privileged & 29.07 & 20.25 \\
\bottomrule\end{tabular}
\end{table}
\begin{table}[!htbp]
\centering\scriptsize
\caption{Top four evaluated configurations: GPT-5.6 Sol. Evaluation-set-selected descriptive maxima; privileged inputs are explicitly marked.}
\label{tab:top4_1}
\begin{tabular}{lrrrrrr}\toprule
Router & Executor & Planner & Policy & Input & EA & R+E \\
\midrule
Hermes & Adapter & Direct & Ask & Privileged & 61.25 & 22.64 \\
Hermes & Adapter & Oracle & Ask & Privileged & 61.25 & 22.64 \\
ReAct & Adapter & Direct & Ask & Privileged & 50.48 & 22.58 \\
ReAct & Adapter & Oracle & Ask & Privileged & 50.48 & 22.58 \\
\bottomrule\end{tabular}
\end{table}
\begin{table}[!htbp]
\centering\scriptsize
\caption{Top four evaluated configurations: DeepSeek-V4-Flash. Evaluation-set-selected descriptive maxima; privileged inputs are explicitly marked.}
\label{tab:top4_2}
\begin{tabular}{lrrrrrr}\toprule
Router & Executor & Planner & Policy & Input & EA & R+E \\
\midrule
Hermes & Adapter & Direct & Ask & Privileged & 58.68 & 22.64 \\
Hermes & Adapter & Oracle & Ask & Privileged & 58.59 & 22.64 \\
ReAct & Adapter & Direct & Ask & Privileged & 26.60 & 19.26 \\
ReAct & Adapter & Oracle & Ask & Privileged & 26.56 & 19.26 \\
\bottomrule\end{tabular}
\end{table}
\begin{table}[!htbp]
\centering\scriptsize
\caption{Top four evaluated configurations: Kimi K3. Evaluation-set-selected descriptive maxima; privileged inputs are explicitly marked.}
\label{tab:top4_3}
\begin{tabular}{lrrrrrr}\toprule
Router & Executor & Planner & Policy & Input & EA & R+E \\
\midrule
ReAct & Adapter & Direct & Ask & Privileged & 47.17 & 23.37 \\
ReAct & Adapter & Oracle & Ask & Privileged & 47.21 & 23.37 \\
Hermes & Adapter & Direct & Ask & Privileged & 59.09 & 22.64 \\
Hermes & Adapter & Oracle & Ask & Privileged & 59.18 & 22.64 \\
\bottomrule\end{tabular}
\end{table}
\FloatBarrier
\begin{table}[!htbp]
\centering\scriptsize
\caption{Clarification mechanisms with LLM planning and non-privileged inputs (percent). Conditional rates divide pooled event counts by pooled Ask counts across fixed cells.}
\label{tab:mechanism_new}
\begin{tabular}{lrrrr}\toprule
Backend & Ask rate & Change/Ask & Repair/Ask & Harm/Ask \\
\midrule
Claude Opus 5 & 100.00 & 58.57 & 45.71 & 2.54 \\
GPT-5.6 Sol & 99.37 & 57.03 & 43.13 & 2.40 \\
DeepSeek-V4-Flash & 97.14 & 54.41 & 43.30 & 1.63 \\
Kimi K3 & 94.60 & 56.54 & 44.63 & 2.52 \\
\bottomrule\end{tabular}
\end{table}
Marginal effects over both planners and all input modes are not interchangeable with the fixed-control comparisons in the main text.